\documentclass[10pt,twocolumn,letterpaper]{article}

\usepackage[pagenumbers]{cvpr} 

\makeatletter
\providecommand{\@LN}[2]{}
\makeatother

\definecolor{cvprblue}{rgb}{0.21,0.49,0.74}
\usepackage[pagebackref,breaklinks,colorlinks,allcolors=cvprblue]{hyperref}
\usepackage{tabularray,xcolor, multirow}
\usepackage{algorithm}
\usepackage{listings}
\usepackage{amsmath}
\usepackage{color, colortbl} 
\UseTblrLibrary{siunitx}

\usepackage[accsupp]{axessibility}  

\definecolor{darkF7E0D5}{RGB}{209,154,128}
\definecolor{Gray}{gray}{0.90}
\definecolor{light}{RGB}{179, 224, 255}
\definecolor{darkblue}{RGB}{0, 112, 188}

\definecolor{newyellow}{RGB}{197, 197, 0}
\newcommand{\rc}{\cellcolor{light}}

\newcommand{\rownumber}[1]{\textcolor{darkblue}{#1}}
\newcommand{\method}{\textcolor{black}{\text{COSMOS}}\xspace}
\newcommand{\supp}{Supplementary Material\xspace}
\newcommand{\myparagraph}[1]{\noindent{\bf{#1}}}
\usepackage{etoolbox}
\usepackage{float}
\makeatletter
\AfterEndEnvironment{algorithm}{\let\@algcomment\relax}
\AtEndEnvironment{algorithm}{\kern2pt\hrule\relax\vskip3pt\@algcomment}
\let\@algcomment\relax
\newcommand\algcomment[1]{\def\@algcomment{\footnotesize#1}}
\renewcommand\fs@ruled{\def\@fs@cfont{\bfseries}\let\@fs@capt\floatc@ruled
  \def\@fs@pre{\hrule height.8pt depth0pt \kern2pt}%
  \def\@fs@post{}%
  \def\@fs@mid{\kern2pt\hrule\kern2pt}%
  \let\@fs@iftopcapt\iftrue}
\makeatother

\def\paperID{8415} 
\def\confName{CVPR}
\def\confYear{2025}

\title{COSMOS: Cross-Modality Self-Distillation for Vision Language Pre-training} 

\author{Sanghwan Kim\textsuperscript{1,2,3,4}, Rui Xiao\textsuperscript{1, 2}, Mariana-Iuliana Georgescu\textsuperscript{1,2,3,4}, \\ Stephan Alaniz\textsuperscript{1,2,3,4}, Zeynep Akata\textsuperscript{1,2,3,4} \\[5pt]
\textsuperscript{1}Technical University of Munich  \quad \textsuperscript{2}Munich Center for Machine Learning (MCML) \\
\textsuperscript{3}Helmholtz Munich \quad \textsuperscript{4}Munich Data Science Institute (MDSI)\\
}

\begin{document}
\maketitle
\begin{abstract}
Vision-Language Models (VLMs) trained with contrastive loss have achieved significant advancements in various vision and language tasks. However, the global nature of the contrastive loss makes VLMs focus predominantly on foreground objects, neglecting other crucial information in the image, which limits their effectiveness in downstream tasks. To address these challenges, we propose \method: \textbf{C}r\textbf{OS}s-\textbf{MO}dality \textbf{S}elf-distillation for vision-language pre-training that integrates a novel text-cropping strategy and cross-attention module into a self-supervised learning framework. We create global and local views of images and texts (i.e., multi-modal augmentations), which are essential for self-distillation in VLMs. We further introduce a cross-attention module, enabling \method to learn comprehensive cross-modal representations optimized via a cross-modality self-distillation loss. \method consistently outperforms previous strong baselines on various zero-shot downstream tasks, including retrieval, classification, and semantic segmentation. Additionally, it surpasses CLIP-based models trained on larger datasets in visual perception and contextual understanding tasks. Code is available at \href{https://github.com/ExplainableML/cosmos}{https://github.com/ExplainableML/cosmos}.

\end{abstract}
    
\section{Introduction}
\label{sec:intro}

CLIP~\cite{radford2021learning} uses an instance-level contrastive loss~\cite{oord2018representation} to learn text and image representations, where the text gets embedded close to its corresponding image. The contrastive loss function employed to train CLIP is inherently global, as it matches the entire image to the text. 
This leads to the dominant objects in the image suppressing the recognition of other smaller objects~\cite{chen2021intriguing, robinson2021can, zhang2024learning, li2023addressing}, which results in poor performance for dense prediction tasks, such as semantic segmentation~\cite{naeem2023silc}, and failure to distinguish similar-looking images with different visual patterns~\cite{tong2024eyes}. CLIP also demonstrates a lack of contextual understanding, as its text encoder perceives a caption as a bag-of-words, ignoring the sequence of words~\cite{thrush2022winoground, yuksekgonul2022and, hsieh2024sugarcrepe}.

\begin{figure}[t]
    \centering
    \includegraphics[width=1.0\linewidth]{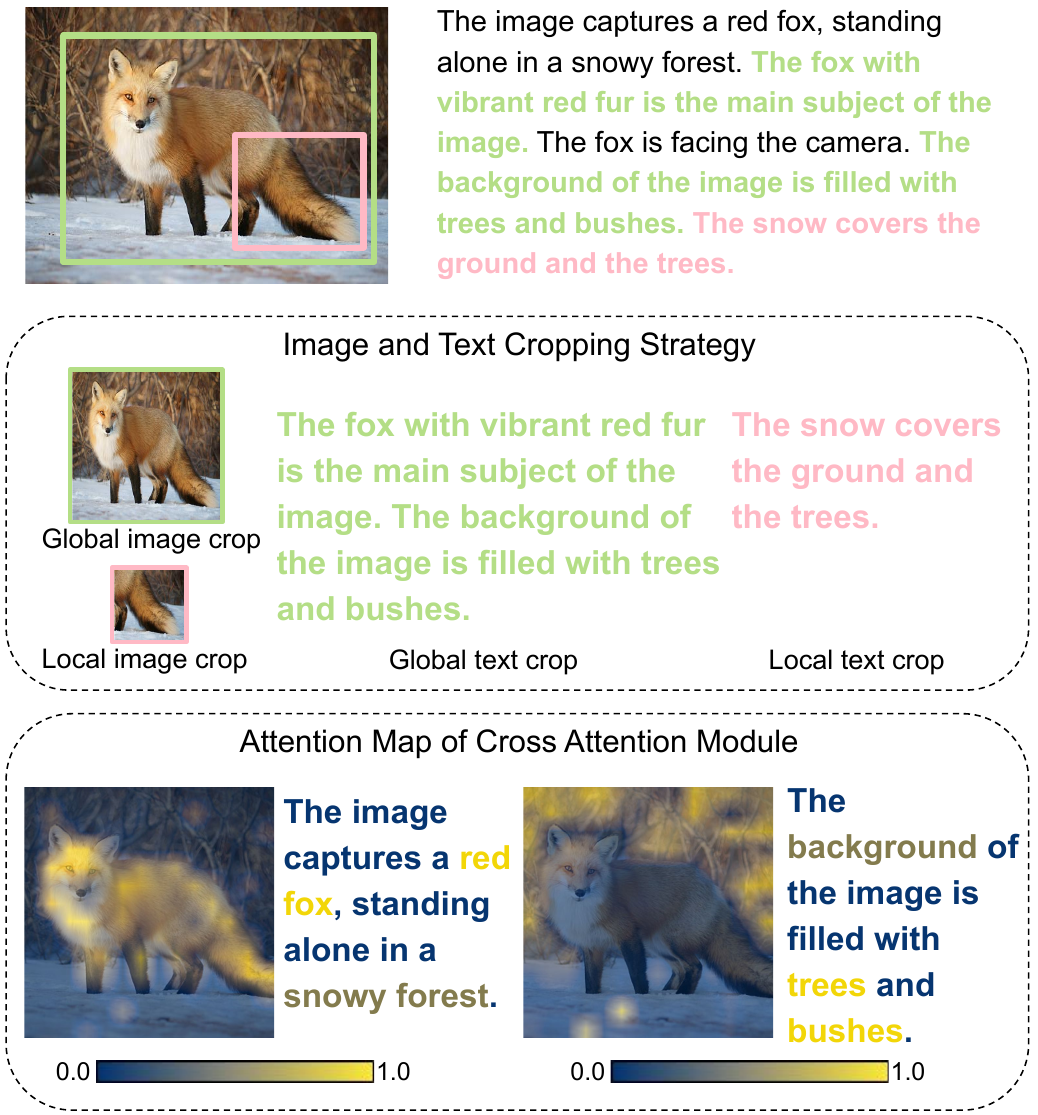}
\caption{We randomly crop the image and randomly select captions (1-5 sentences) to build global and local crops (top). In our cross attention module, two modalities are conditioned on each other to create attention maps (normalized to 0-1, bottom). 
}
    \label{fig:teaser}
\end{figure}

Previous works~\cite{li2021supervision, mu2022slip, naeem2023silc, dong2023maskclip, sameni2024building} have tackled these limitations by masking the text or the image~\cite{dong2023maskclip, li2021supervision, sameni2024building} and explicitly optimizing local-to-global image representation. For instance, SLIP~\cite{mu2022slip}  and SILC~\cite{naeem2023silc} improved the visual representations by integrating self-supervision into the CLIP image encoder. However, we argue that enhancing only the image encoder leads to sub-optimal results, since this does not fully exploit the multi-modality of the data. Therefore, we propose \method, to learn \textbf{C}r\textbf{OS}s-\textbf{MO}dality embeddings through \textbf{S}elf-distillation. Unlike previous self-supervised methods~\cite{mu2022slip,naeem2023silc,dong2023maskclip}, \method incorporates a unique text-cropping strategy and a cross-attention module to learn multi-modal representations for various downstream tasks. 

Our text-cropping strategy (\cref{fig:teaser}, top) is inspired by the multi-crop augmentation for images~\cite{caron2021emerging}, which optimizes local-to-global image correspondence during training. We randomly sample from the long synthetic caption datasets~\cite{zheng2025dreamlip,singla2024pixels} to construct sentences of varying length. By applying multi-crop augmentation to both images and text, we enhance the visual and text encoders simultaneously within a self-distillation framework~\cite{caron2021emerging}, aligning the representation of the student VLM with that of the teacher. 

Our cross-attention module, integrated into the student, allows the model to attend to the other modality's representation, ensuring that cross-modality information is captured during training. Inside the module, image patch tokens are conditioned on texts, or word tokens are conditioned on images, to learn visual and textual grounding. As illustrated in \cref{fig:teaser} (bottom), our cross-attention module focuses on different objects from the entire image, which are highlighted in both images and captions based on the attention weights. This demonstrates that \method is able to effectively localize relevant information in images and captions. 

We demonstrate the adaptability of \method by pre-training it on 4 datasets varying from 3M to 30M pairs. To evaluate our method, we perform zero-shot experiments on 2 retrieval, 11 classification, and 6 segmentation benchmarks, as well as, the visual perception~\cite{tong2024eyes} and contextual understanding tasks~\cite{hsieh2024sugarcrepe, hendricks2021probing}. The extensive experiments on these downstream tasks reveal that our method consistently outperforms the previous CLIP-based methods, under the same pre-training data budget. Moreover, \method pre-trained on only 12 million examples outperforms the OpenCLIP~\cite{cherti2023reproducible} models trained on billions of image-text pairs, in both image-to-text and text-to-image retrieval tasks.

Our contributions are summarized as follows:
(1) We design a text-cropping strategy which promotes local-to-global text representation learning, fully leveraging multi-modality in the self-distillation of VLMs.
(2) We demonstrate that cross-modality embeddings are effective to simultaneously self-distill both image and text encoders.
(3) \method reveals fine-grained perception of images and compositional understanding of language, outperforming baselines trained on larger datasets. 

\section{Related Work}
\label{sec:related}

\begin{figure*}[h!]
    \centering
    \includegraphics[width=1.0\linewidth]{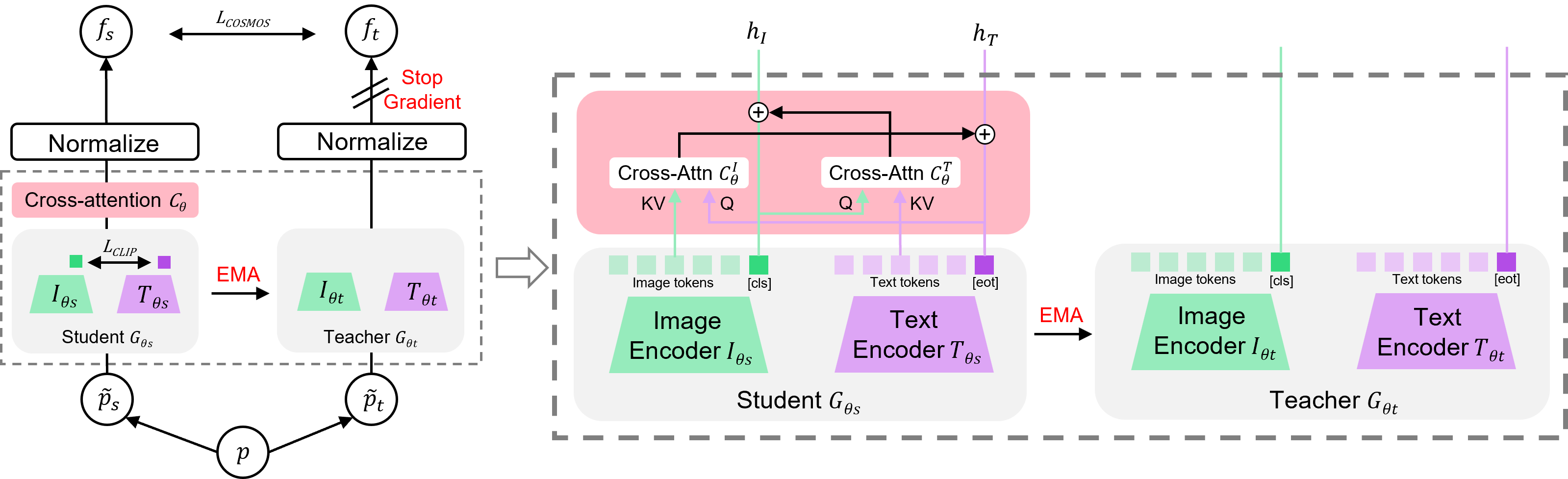}
    \caption{
        \textbf{An overview of \method.} Left: Our VLM pre-training mechanism is based on the student-teacher framework with contrastive loss ($\mathcal{L}_{\text{CLIP}}$) for multi-modal alignment, and cross-modality self-distillation loss ($\mathcal{L}_{\text{COSMOS}}$) for fine-grained representation learning. Right: The architecture of the student and teacher model with cross-attention modules that extract cross-modality information from the student. 
        } 
    \label{fig:framework}
    \vspace{-3mm}
\end{figure*}

Self-Supervised Learning (SSL) learns discriminative features from unlabeled data~\cite{gui2024survey}. Initially, models like BERT~\cite{devlin2018bert} and GPT~\cite{radford2019language} were designed to predict masked words or the next word in a sequence, eliminating the need for additional labels. Similarly, vision models benefit from self-supervised techniques, either with contrastive learning or masked image modeling. Based on contrastive learning, BYOL~\cite{grill2020bootstrap} and DINO~\cite{caron2021emerging} trained a student network to produce image embeddings that closely resemble those of a teacher network, with the teacher being an exponential moving average of the student. On the other hand, masked image modeling divides the input image into visual tokens and predicts a randomly masked subset of them. For example, MAE~\cite{he2022masked} directly predicted the original pixel from the masked patches, and BEiT~\cite{bao2021beit} employed a pre-trained tokenizer as its target.

Contrastive learning in vision-language pre-training has achieved great success in various downstream tasks, demonstrating strong zero-shot transfer abilities~\cite{radford2021learning, jia2021scaling, yao2021filip, gao2022pyramidclip, geng2023hiclip, lee2022uniclip, gao2023softclip}. As these models are trained to learn visual representations aligned with the language representation, most Multi-modal Large Language Models (MLLMs)~\cite{alayrac2022flamingo, li2023blip, liu2024visual, liu2024improved, zhu2023minigpt} adopt these vision encoders to process the image input. However, instance-level contrastive learning suffers from feature suppression, where the model learns only the dominant features in the data while neglecting other valuable features~\cite{chen2021intriguing, robinson2021can, zhang2024learning, li2023addressing, tamkin2024feature, xue2023features, li2023scaling, tong2024eyes}. In other words, the model creates so-called simple shortcut features and decision rules that do not consider all available distinguishing features. Some works~\cite{chen2023sharegpt4v, singla2024pixels, urbanek2024picture, zhang2025long, onoe2024docci, zheng2025dreamlip, wu2024lotlip} have employed datasets equipped with long synthetic captions generated by MLLMs. These datasets help mitigate feature suppression by learning different features described in the captions and avoiding overfitting to a single caption. Referring to \citet{zheng2025dreamlip}, we employ image datasets with long synthetic captions. Instead of simply training our model with these additional captions, we create text augmentation using the synthetic captions, and introduce the cross-attention module for self-distillation. 

Following the success of SSL, recent works~\cite{li2021supervision, mu2022slip, naeem2023silc, dong2023maskclip, sameni2024building, kim2024expediting} apply SSL objectives to contrastive vision-language pre-training. These approaches also tackle the feature suppression limitation, where the new objective reduces the inductive bias to prevent the creation of shortcut features. For example, SLIP~\cite{mu2022slip} combined the SimCLR~\cite{chen2020simple} loss with the contrastive loss, resulting in better transferability compared to the original CLIP. Similarly, SILC~\cite{naeem2023silc} explicitly matched local crops to global crops during training inspired by DINO~\cite{caron2021emerging}, while MaskCLIP~\cite{dong2023maskclip} learned more fine-grained features by applying masked modeling to both images and text. Similar to SLIP~\cite{mu2022slip} and SILC~\cite{naeem2023silc}, we also employ self-supervised learning for vision-language pre-training. While SLIP~\cite{mu2022slip} and SILC~\cite{naeem2023silc} only improve the image representation without improving the text representation, we propose a text-cropping technique that enables us to improve the text and image representations simultaneously with our novel cross-modality self-distillation loss.

\section{COSMOS Model}
\label{sec:method}

We describe our augmentation technique, the key component in self-distillation, in~\cref{sub:aug}, \method's architecture in~\cref{sub:arch}, and the training objective in~\cref{sub:obj}.

\subsection{Text and Image Augmentations}
\label{sub:aug}

\myparagraph{Text Cropping Strategy.} To perform self-supervised learning based on text, we propose a text-cropping strategy utilizing long caption datasets~\cite{zheng2025dreamlip, singla2024pixels}. These datasets consist of images paired with detailed synthetic captions generated by MLLMs. Inspired by the multi-crop strategy for images~\cite{caron2021emerging}, we define \textit{global} and \textit{local} concepts for text captions. The \textit{global} crop of a caption comprises one to five sentences randomly sampled from the long synthetic caption, while the \textit{local} crop of a caption consists of a single sentence, also randomly sampled from the long caption. The number of sentences in the global captions is randomly determined at each training iteration. This adaptation ensures that the global caption generally describes larger areas of the original image, e.g. \textit{the fox with vibrant red fur is the main subject of the image}, in \cref{fig:teaser} top, while the local caption focuses on relatively smaller regions, e.g. \textit{the snow covers the ground and the trees}.

\myparagraph{Image Cropping Strategy.} Following \cite{caron2021emerging}, we adopt the standard multi-crop strategy~\cite{caron2020unsupervised} for images as illustrated in \cref{fig:teaser} top, utilizing two global views and several local views. Both local and global views are processed through the student encoder, while only the global views are passed through the teacher encoder, thereby promoting \textit{local-to-global} correspondences. It is noteworthy that the image cropping and text cropping strategies are executed independently, meaning their global and local crops might not contain the same information. For instance, in \cref{fig:teaser} top, the local image crop displays the fox's tail, while the local text crop describes the image's background. This configuration is the standard setting of \method, unless otherwise stated.

\subsection{Model Architecture}
\label{sub:arch} 
The overall architecture of \method is depicted in \cref{fig:framework}. Both the student model $\textbf{G}_{\theta_s}$ and the teacher model $\textbf{G}_{\theta_t}$ are Vision-Language Models (VLMs), each comprising an image encoder $\textbf{I}_{\theta}$ and a text encoder $\textbf{T}_{\theta}$, such that $\textbf{G}_{\theta_*} = \{\textbf{I}_{\theta_*}, \textbf{T}_{\theta_*}\}$, where $* \in \{s, t\}$. The student and teacher models share the same architecture and are initialized with identical weights. Additionally, the student model includes a cross-attention module $\textbf{C}_\theta = \{\textbf{C}^{I}_\theta, \textbf{C}^{T}_\theta\}$, which extracts cross-modal embeddings.

As shown on the left side of \cref{fig:framework}, two random multi-modal transformations are applied to an image-text pair $p$, resulting in $\Tilde{p}_s$ and $\Tilde{p}_t$. The student model $\textbf{G}_{\theta_s}$ processes $\Tilde{p}_s$, followed by the cross-attention module and normalization operation to produce the output $f_s$. Similarly, the teacher model $\textbf{G}_{\theta_t}$ processes $\Tilde{p}_t$ to generate the output $f_t$, which serves as the target. The student model, along with the cross-attention module, is optimized by matching its output $f_s$ with the teacher's output $f_t$ ($\mathcal{L}_{\text{COSMOS}}$), while standard contrastive loss is calculated within the student ($\mathcal{L}_{\text{CLIP}}$). A stop gradient operation is applied to the teacher model to ensure that gradients propagate only through the student. The teacher parameters $\theta_t$ are updated at each iteration using the exponential moving average (EMA) of the student parameters $\theta_s$ (i.e., $\theta_t = \lambda\theta_t + (1-\lambda)\theta_s$ with the parameter $\lambda$  defining the update schedule). This learning mechanism, combined with multi-modal augmentations, enables \method to learn the correspondence between local features from the student and global patterns from the teacher.

The VLMs and the cross-attention module are detailed on the right side of \cref{fig:framework}. Student image and text embeddings are extracted from the image encoder $\textbf{I}_{\theta_s}$ and text encoder $\textbf{T}_{\theta_s}$ respectively. The image embedding consists of image tokens and a class token $[\text{cls}]$, while the text embedding consists of text tokens and an end-of-text token $[\text{eot}]$. The final cross-modal embeddings of image $h_{I}$ and text $h_{T}$ are calculated as follows: 
\begin{align}
        h_{I} &= \textbf{C}^{T}_\theta(q=[\text{cls}], kv=\text{txt-tok}) + [\text{cls}] \label{eq:img_cross_emb}\\
        h_{T} &= \textbf{C}^{I}_\theta(q=[\text{eot}], kv=\text{img-tok}) + [\text{eot}] \label{eq:txt_cross_emb}
\end{align}
In \cref{eq:img_cross_emb}, $\textbf{C}_\theta^\text{T}$ receives the $[\text{cls}]$ token as a query and text tokens (txt-tok) as keys and values and then, the output of $\textbf{C}_\theta^\text{T}$ is integrated into the $[\text{cls}]$ token to obtain the image cross-modal embedding $h_{I}$. Similarly in \cref{eq:txt_cross_emb}, the text cross-modal embedding $h_{T}$ is created by querying the $[\text{eot}]$ token in the cross-attention module $\mathcal{C}_\theta^\text{I}$, with image tokens (img-tok) as keys and values, and adding it to the $[\text{eot}]$ token. This process effectively combines information from the relevant text (or image) tokens associated with the $[\text{cls}]$ (or $[\text{eot}]$) token. Consequently, the distillation signal flows into both encoders during training, encouraging our model to learn both visual and textual grounding. In practice, $\textbf{C}_\theta^\text{T}$ and $\textbf{C}_\theta^\text{I}$ utilize image and text tokens from the global crop of images and captions as keys and values, facilitating a fluent pooling of information through the cross-attention mechanism. As illustrated in \cref{fig:teaser} bottom and \cref{fig:qualitative}, our cross-attention module effectively attends to various parts of images and texts, enabling the generation of fine-grained embeddings.
 
\subsection{Training Objective}
\label{sub:obj}
Similar to CLIP~\cite{radford2021learning}, we employ a standard contrastive loss to align the image and text embeddings produced by the student model $\textbf{G}_{\theta_s}$. For each batch of size $B$, the image and text encoders, $\textbf{I}_{\theta_s}$ and $\textbf{T}_{\theta_s}$, extract sets of $[\text{cls}]$ tokens ($\text{CLS}_s$) and $[\text{eot}]$ tokens ($\text{EOT}_s$). We then apply a symmetric InfoNCE loss formulation~\cite{oord2018representation}, where paired image-text embeddings form positive pairs, while unpaired embeddings are treated as negative samples. The contrastive InfoNCE loss is defined as:
\begin{equation}
    \mathcal{L}_\text{nce}(x,y) = -\frac{1}{B} \sum_{i=1}^B \log \frac{ \exp \left({ \langle x_i, y_i \rangle/\tau}\right)}{\sum_{j=1}^{B}  \exp \left({\langle x_j, y_i \rangle/\tau}\right)}
    \label{eq.con}    
\end{equation}
where $\langle \cdot, \cdot \rangle$ denotes the cosine similarity function, $x$ and $y$ are embeddings, and $\tau$ is a learnable temperature parameter. The symmetric InfoNCE loss is composed of $y$-to-$x$ loss and $x$-to-$y$ loss, and is defined as:
\begin{equation}
   \mathcal{L}_{\text{sym-nce}}(x,y) =\frac{1}{2}(\mathcal{L}_\text{nce}(x,y) + \mathcal{L}_\text{nce}(y,x)). 
\end{equation}
The total contrastive loss $\mathcal{L}_{\text{CLIP}}$ is calculated between the image and text embeddings of the student as follows.
\begin{equation}
   \mathcal{L}_{\text{CLIP}} = \mathcal{L}_{\text{sym-nce}}(\text{CLS}_s, \text{EOT}_s).
\end{equation}

Next, we compute the cross-modality self-distillation (COSMOS) loss between the cross-modal tokens produced by the student and the $[\text{cls}]$ and $[\text{eot}]$ tokens produced by the teacher. Given the image cross-modal tokens $h_{I}$  and the text cross-modal tokens $ h_{T}$ computed following \cref{eq:img_cross_emb} and \cref{eq:txt_cross_emb}, the COSMOS loss is formulated as:
\begin{multline}
 \mathcal{L}_{\text{COSMOS}} = \frac{1}{4} ( 
 \mathcal{L}_\text{sym-nce}(h_{I}, \text{CLS}_t)    +  \mathcal{L}_\text{sym-nce}(h_{I}, \text{EOT}_t) \\   + \mathcal{L}_\text{sym-nce}(h_{T}, \text{CLS}_t)   +  \mathcal{L}_\text{sym-nce}(h_{T}, \text{EOT}_t))
\end{multline}
where $\text{CLS}_t$ and  $\text{EOT}_t$ are the set of $[\text{cls}]$ and $[\text{eot}]$ tokens extracted by the teacher $\textbf{G}_{\theta_t}$.
The final training loss $\mathcal{L}_{\text{total}} $ is the sum of the CLIP loss and the COSMOS loss:
\begin{equation}
\mathcal{L}_{\text{total}} = \mathcal{L}_{\text{CLIP}} + \mathcal{L}_{\text{COSMOS}}.
\end{equation}
As mentioned in \cref{sub:aug}, all crops are passed through the student while only global crops are given to the teacher. Thus, $\mathcal{L}_{\text{CLIP}}$ is calculated between all crops within the student, optimizing the model to construct the overall structure of image and text representations by aligning the embeddings through contrastive learning. On the other hand, $\mathcal{L}_{\text{COSMOS}}$ is obtained between global crops given to the teacher and all crops given to the student, encouraging the model to learn fine-grained representations by predicting the rich global context of the teacher from the local feature of the student (the detailed pseudocode of the training objective is in \supp \cref{alg:code}).

\begin{table*}[h!]
\centering
\setlength\tabcolsep{1.0pt}
\resizebox{.87\linewidth}{!}{

\begin{tabular}
{l<{\hspace{3.0em}}c<{\hspace{1.0em}} | c<{\hspace{0.7em}}c<{\hspace{0.7em}}c<{\hspace{0.7em}} | c<{\hspace{0.7em}}c<{\hspace{0.7em}}c<{\hspace{0.7em}} | c<{\hspace{0.7em}}c<{\hspace{0.7em}}c<{\hspace{0.7em}} | c<{\hspace{0.7em}}c<{\hspace{0.7em}}c<{\hspace{0.7em}}}

\toprule
&&\multicolumn{6}{c|}{Flickr30K}& \multicolumn{6}{c}{MSCOCO}\\
&&\multicolumn{3}{c|}{Image-to-text} & \multicolumn{3}{c|}{Text-to-image} & \multicolumn{3}{c|}{Image-to-text} & \multicolumn{3}{c}{Text-to-image} \\
Data & Method & R@1 & R@5 & R@10 & R@1 & R@5 & R@10 & R@1 & R@5 & R@10 & R@1 & R@5 & R@10 \\
\midrule
\multirow{5}{3em}{CC3M}
& MLLM-A$^{*}$~\cite{liu2023mllms} & 63.5 & 86.6 & 91.7 & 49.3 & 74.8 & 83.1 & 35.9 & 62.4 & 73.9 & 26.5 & 51.1 & 62.7 \\
\cmidrule(lr){2-14} 
& CLIP~\cite{radford2021learning} & 68.4 & 88.9 & 93.2 & 52.1 & 78.4 & 85.5 & 40.2 & 67.0 & 77.5 & 27.2 & 53.5 & 65.0  \\ 
& SigLIP~\cite{zhai2023sigmoid}  & 69.8 & 90.4 & 94.5 & 53.1 & 78.5 & 86.1 & 40.1 & 67.5 & 78.1 & 28.3 & 54.4 & 65.9  \\ 
& DreamLIP$^{\dag}$~\cite{zheng2025dreamlip}  & 69.2 & 91.5 & 95.9 & 53.5 & 78.3 & 85.5 & 40.8 & 68.5 & 79.4 & 29.8 & 55.4 & 66.3 \\ 
& \rc\method    & \rc\textbf{84.1} & \rc\textbf{96.2} & \rc\textbf{98.9} & \rc\textbf{68.6} & \rc\textbf{88.7} & \rc\textbf{93.4} & \rc\textbf{53.1} & \rc\textbf{78.3} & \rc\textbf{86.7} & \rc\textbf{40.1} & \rc\textbf{66.8} & \rc\textbf{77.1} \\ 
\midrule
\multirow{7}{3em}{CC12M}
& SLIP$^{*}$~\cite{mu2022slip}              & 62.5& 87.2& 92.1& 46.6& 73.3& 80.9& 37.6& 64.9& 75.5& 26.8& 51.4& 62.7\\ 
& LaCLIP$^{*}$~\cite{fan2024improving}      & 63.9& 86.5& 92.6& 51.6& 78.8& 86.2& 38.0& 64.8& 75.0& 26.5& 51.2& 62.6 \\ 
& LaSF-CLIP$^{*}$~\cite{sameni2024building} & 71.8& 91.9& 95.2& 59.9& 84.2& 90.9& 44.3& 71.3& 80.2& 31.4& 57.3& 68.1 \\ 
\cmidrule(lr){2-14} 
& CLIP~\cite{radford2021learning}   & 81.8 & 96.5 & 98.1 & 66.9 & 87.7 & 92.7 & 56.2 & 80.4 & 88.0 & 39.8 & 66.4 & 76.6 \\ 
& SigLIP~\cite{zhai2023sigmoid}  & 82.5 & 96.1 & 97.8 & 66.7 & 88.1 & 92.8 & 55.4 & 79.7 & 87.8 & 40.4 & 67.0 & 76.6 \\ 
& DreamLIP$^{\dag}$~\cite{zheng2025dreamlip}  & 84.1 & 97.8 & \textbf{99.0} & 68.3 & 89.4 & 93.4 & 54.0 & 78.2 & 86.4 & 40.7 & 66.5 & 76.4 \\ 
& \rc\method & \rc\textbf{91.4} & \rc\textbf{98.2} & \rc98.8 & \rc\textbf{76.2} & \rc\textbf{92.9} & \rc\textbf{96.3} & \rc\textbf{64.2} & \rc\textbf{85.6} & \rc\textbf{91.6} & \rc\textbf{48.9} & \rc\textbf{74.3} & \rc\textbf{82.9} \\ 
\midrule
\multirow{7}{3em}{YFCC15M}
& SLIP$^{*}$~\cite{mu2022slip}           & 58.6 & 85.1 & 91.7  &41.3 & 68.7 & 78.6 & 33.4 & 59.8 & 70.6 & 21.5 & 44.4 & 56.3 \\ 
& MaskCLIP$^{*}$~\cite{dong2023maskclip} & 70.1 & 90.3 & 95.3 & 45.6 & 73.4 & 82.1 & 41.4 & 67.9 & 77.5 & 25.5 & 49.7& 61.3 \\ 
& SF-CLIP$^{*}$~\cite{sameni2024building} & 68.7 &90.4 &94.8 &46.2& 73.2& 82.7 &41.8 &68.3 &78.4 &25.5 &50.2 &61.3 \\  
\cmidrule(lr){2-14} 
& CLIP~\cite{radford2021learning}   & 89.1 & 97.6 & 98.8 & 72.4 & 90.7 & 94.9 & 61.0 & 85.0 & 90.9 & 44.7 & 71.2 & 80.3 \\ 
& SigLIP~\cite{zhai2023sigmoid}  & 89.9 & 97.5 & 98.9 & 73.6 & 92.1 & 95.5 & 62.5 & 85.3 & 91.8 & 46.5 & 72.8 & 81.9 \\ 
& DreamLIP$^{\dag}$~\cite{zheng2025dreamlip}   & 87.3 & 98.1 & 99.0 & 70.0 & 89.2 & 93.4 & 57.0 & 80.9 & 88.4 & 42.4 & 68.5 & 78.2 \\ 
& \rc\method          & \rc\textbf{92.6} & \rc\textbf{99.1} & \rc\textbf{99.7} & \rc\textbf{79.6} & \rc\textbf{94.2} & \rc\textbf{96.8} & \rc\textbf{67.5} & \rc\textbf{88.2} & \rc\textbf{93.5} & \rc\textbf{50.9} & \rc\textbf{76.0} & \rc\textbf{84.2} \\ 
\midrule
\multirow{4}{3em}{\text{Merged-30M}}
& CLIP~\cite{radford2021learning}  & 90.5 & 98.7 & 99.4 & 75.9 & 93.5 & 96.7 & 63.2 & 85.6 & 91.7 & 48.2 & 73.9 & 82.4 \\  
& SigLIP~\cite{zhai2023sigmoid}  & 90.5 & 98.3 & 99.4 & 76.7 & 93.2 & 96.1 & 64.1 & 86.2 & 91.9 & 49.0 & 74.5 & 83.2 \\ 
& DreamLIP$^{\dag}$~\cite{zheng2025dreamlip}  & 89.9 & 99.0 & 99.6 & 73.3 & 91.9 & 95.7 & 62.3 & 84.5 & 90.3 & 44.9 & 69.9 & 79.2\\ 
& \rc\method          & \rc\textbf{92.9} & \rc\textbf{99.4} & \rc\textbf{99.9} & \rc\textbf{80.3} & \rc\textbf{95.3} & \rc\textbf{97.6} & \rc\textbf{68.0} & \rc\textbf{87.8} & \rc\textbf{92.5} & \rc\textbf{52.5} & \rc\textbf{77.2} & \rc\textbf{84.9} \\ 
\midrule
\multirow{1}{3em}{\text{LAION-400M}}
& OpenCLIP$^{\dag}$~\cite{cherti2023reproducible}  & 86.0 & 97.3 & 98.9 & 68.1 & 88.6 & 93.3 & 56.5 & 80.4 & 87.3 & 37.9 & 63.2 & 73.3 \\
\midrule
\multirow{1}{3em}{\text{DataComp-1B}}
& OpenCLIP$^{\dag}$~\cite{cherti2023reproducible}  & 84.5 & 96.8 & 98.4 & 67.9 & 88.0 & 93.0 & 58.2 & 80.8 & 88.5 & 39.8 & 65.5 & 75.0 \\
\midrule
\multirow{1}{3em}{\text{LAION-2B}}
& OpenCLIP$^{\dag}$~\cite{cherti2023reproducible}  & 87.6 & 97.7 & 99.5 & 71.9 & 90.5 & 94.8 & 59.3 & 82.5 & 89.0 & 41.7 & 67.1 & 76.5 \\
\midrule
\multirow{2}{3em}{\text{2.5B}}
& MetaCLIP$^{*}$~\cite{xu2023demystifying}  & 85.9 & 97.3 & 98.9 & 70.5 & 90.7 & 94.6 & 59.4 & 80.6 & 87.9 & 41.4 & 67.2 & 77.0 \\
& Llip$^{*}$~\cite{lavoie2024modeling}  & 90.1 & 98.5 & 99.6 & 75.1 & 92.8 & 96.2 & 63.4 & 84.3 & 90.3 & 45.6 & 70.8 & 79.7 \\
\bottomrule         
\end{tabular}}
\caption{\textbf{Zero-shot image-text retrieval results} in terms of R@1, R@5, and R@10 on the Flickr30K~\cite{young2014image} and MSCOCO~\cite{lin2014microsoft} datasets. The vision encoder architecture is ViT-B/16. The best results are highlighted in \textbf{bold}. Results are reproduced with our setup for fair comparison unless otherwise marked. $^{*}$: Results copied from their work. $^{\dag}$: Results obtained using their official pre-trained weights.}
\label{tab:zs_ret}
\vspace{-3mm}
\end{table*}

\noindent
\textbf{Simple Loss Formulation.} Previous works~\cite{li2021supervision, naeem2023silc, dong2023maskclip, sameni2024building} often rescale the CLIP loss when combining it with distillation objectives, to ensure that the new loss does not disrupt the original learning mechanism of CLIP. Simply adding self-distillation loss without scaling the CLIP loss often underperforms the original model.  To overcome this limitation, one must grid-search for the optimal loss scale factor. In our method, both CLIP and the COSMOS loss are updated at the same scale, eliminating the need for additional hyperparameter tuning to adjust the ratio of the two losses.

\section{Experiments}
\label{sec:experiment}

\begin{table*}[h!]
    \setlength\tabcolsep{1.0pt}
    \centering
    \resizebox{0.77\linewidth}{!}{
    \begin{tabular}{c<{\hspace{3.0em}}c<{\hspace{1.0em}}|c<{\hspace{0.7em}}c<{\hspace{0.7em}}c<{\hspace{0.7em}}c<{\hspace{0.7em}}c<{\hspace{0.7em}}c<{\hspace{0.7em}}c<{\hspace{0.7em}}c<{\hspace{0.7em}}c<{\hspace{0.7em}}c<{\hspace{0.7em}}c<{\hspace{0.7em}}|c<{\hspace{0.7em}}}
        \toprule[1.2pt]
        Data&Model&
        \rotatebox[origin=lb]{90}{\smash{ Food-101}} & \rotatebox[origin=lb]{90}{\smash{ CIFAR-10}} & \rotatebox[origin=lb]{90}{\smash{ CIFAR-100}}   & \rotatebox[origin=lb]{90}{\smash{ SUN397}}   & \rotatebox[origin=lb]{90}{\smash{ Cars}}     & \rotatebox[origin=lb]{90}{\smash{ Aircraft}}    & \rotatebox[origin=lb]{90}{\smash{ DTD}}      & \rotatebox[origin=lb]{90}{\smash{ Pets}}     & \rotatebox[origin=lb]{90}{\smash{ Caltech-101}} &
        \rotatebox[origin=lb]{90}{\smash{ Flowers}}  & \rotatebox[origin=lb]{90}{\smash{ ImageNet}}  & \rotatebox[origin=lb]{90}{\smash{ Average}} \\
        \midrule
        \multirow{6}{3em}{\rotatebox[origin=c]{0}{ CC3M}} 
        & LaCLIP$^{*}$~\cite{fan2024improving}    & 14.2 & 57.1 & 27.5 & 35.1 & 1.6 & 1.6 & 16.6 & 15.6 & 52.7 & 14.7 & 21.5 & 23.5 \\ 
        & MLLM-A$^{*}$~\cite{liu2023mllms}        & 18.7 & 58.4 & 32.4 & 43.8 & 3.9 & 1.5 & 20.2 & 32.1 & 63.5 & 17.5 & 25.0 & 28.8 \\ 
        \cmidrule(lr){2-14} 
        & CLIP~\cite{radford2021learning}   & 17.9 & 75.0 & 40.8 & 43.1 & 2.6 & 1.0 & 15.3 & 22.1 & 68.9 & 12.6 & 23.9 & 29.4 \\
        & SigLIP~\cite{zhai2023sigmoid}     & 18.4 & 76.4 & 41.9 & 46.9 & 3.0 & 1.4 & 17.6 & 20.6 & 70.4 & 10.8 & 25.9 & 30.3 \\
        & DreamLIP$^{\dag}$~\cite{zheng2025dreamlip} & 23.6 & 75.7 & 44.2 & 46.1 & 3.6 & 1.6 & 18.7 & 27.4 & 73.6 & 18.5 & 31.6 & 33.1 \\
        & \rc\method         & \rc\textbf{31.2} & \rc\textbf{79.1} & \rc\textbf{51.6} & \rc\textbf{54.2} & \rc\textbf{5.2} & \rc\textbf{2.3} & \rc\textbf{27.4} & \rc\textbf{37.5} & \rc\textbf{78.1} & \rc\textbf{25.6} & \rc\textbf{37.1} & \rc\textbf{39.0} \\ 
        \midrule
        \multirow{7}{3em}{\rotatebox[origin=c]{0}{  CC12M}} 
        & LaCLIP$^{*}$~\cite{fan2024improving}    & 60.7 & 75.1 & 43.9 & 57.0 & \textbf{36.3} & \textbf{5.6} & 31.0 & \textbf{72.4} & 83.3 & \textbf{39.9} & 48.4 & 50.3 \\ 
        & LaSLIP$^{*}$~\cite{fan2024improving}    & 62.9 & 82.0 & 50.2 & 59.6 & 32.2 & 4.4 & 30.1 & 70.6 & 82.4 & 37.4 & 49.7 & 51.0 \\ 
        & MLLM-A$^{*}$~\cite{liu2023mllms}        & 60.9 & 83.0 & 55.4 & 59.4 & 24.1 & 3.2 & 30.7 & 64.8 & 79.3 & 36.0 & 47.5 & 49.5 \\ 
        \cmidrule(lr){2-14} 
        &  CLIP~\cite{radford2021learning} & 38.4 & 90.4 & 58.6 & 57.7 & 12.2 & 2.0 & 24.6 & 40.9 & 80.8 & 12.1 & 39.1 & 41.5 \\
        & SigLIP~\cite{zhai2023sigmoid} & 42.3 & 88.9 & 59.6 & 57.6 & 11.2 & 2.3 & 24.1 & 38.4 & 80.0 & 13.5 & 38.9 & 41.5 \\
        & DreamLIP$^{\dag}$~\cite{zheng2025dreamlip} & 59.2 & 87.7 & 63.8 & 54.5 & 29.9 & 5.1 & 29.5 & 60.8 & 82.5 & 32.5 & 50.2 & 50.5 \\
        & \rc\method    & \rc\textbf{63.1} & \rc\textbf{92.8} & \rc\textbf{70.8} & \rc\textbf{64.2} & \rc30.4 & \rc3.4 & \rc\textbf{36.5} & \rc56.9 & \rc\textbf{85.5} & \rc36.3 &\rc\textbf{51.4} & \rc\textbf{53.8} \\ 
        \midrule
        \multirow{7}{3em}{\rotatebox[origin=c]{0}{ YFCC15M}} 
        & SLIP$^{*}$~\cite{mu2022slip}            & 59.9 & 82.6 & 48.6 & -    & 5.4 & 5.6  & 26.6 & 31.4 & 70.9 & 56.3 & 42.8 & - \\ 
        & MaskCLIP$^{*}$~\cite{dong2023maskclip}  & 64.9 & 80.2 & 57.5 & -    & 6.7 & 6.1  & 27.9 & 34.3 & 72.0 & 57.0 & 44.5 & - \\ 
        & SF-CLIP$^{*}$~\cite{sameni2024building} & \textbf{65.0} & 85.0 & 53.6 & -    & 8.2 & \textbf{11.0} & \textbf{35.2} & 38.1 & 72.2 & \textbf{59.5} & 45.2 & - \\ 
        \cmidrule(lr){2-14} 
        & CLIP~\cite{radford2021learning} & 43.5 & 88.0 & 56.8 & 55.8 & 9.3 & 4.7 & 26.0 & 27.9 & 76.7 & 37.0 & 41.7 & 42.5 \\
        & SigLIP~\cite{zhai2023sigmoid} & 45.7 & 87.8 & 58.7 & 58.6 & 9.9 & 4.7 & 28.1 & 32.3 & 79.1 & 36.4 & 44.4 & 44.2 \\
        & DreamLIP$^{\dag}$~\cite{zheng2025dreamlip} & 45.0 & 88.9 & 62.8 & 57.5 & 9.4 & 5.8 & 32.0 & 33.6 & 81.5 & 44.4 & 48.2 & 46.3 \\
        & \rc\method   & \rc55.8 & \rc\textbf{90.4} & \rc\textbf{65.6} & \rc\textbf{59.7} & \rc\textbf{16.8} & \rc5.8 & \rc33.7 & \rc\textbf{39.3} & \rc\textbf{83.1} & \rc46.9 & \rc\textbf{52.4} & \rc\textbf{50.0} \\ 
        \midrule
        \multirow{4}{3em}{\rotatebox[origin=c]{0}{Merged-30M}}  
        &  CLIP~\cite{radford2021learning} & 61.3 & 92.2 & 66.9 & 62.2 & 19.3 & 5.7 & 30.9 & 49.3 & 83.7 & 43.4 & 50.0 & 51.4 \\
        & SigLIP~\cite{zhai2023sigmoid}    & 64.2 & 91.0 & 67.6 & 64.0 & 22.0 & 5.7 & 33.5 & 53.3 & 84.3 & 43.6 & 51.0 & 52.7 \\
        & DreamLIP$^{\dag}$~\cite{zheng2025dreamlip} & \textbf{76.4} & 92.7 & 71.7 & 64.7 & 23.4 & \textbf{8.3} & 33.9 & \textbf{66.3} & 87.6 & \textbf{59.1} & \textbf{58.4} & 58.4 \\
        &   \rc\method    & \rc73.9 & \rc\textbf{93.4} & \rc\textbf{72.3} & \rc\textbf{67.8} & \rc\textbf{31.2} & \rc5.9 & \rc\textbf{37.3} & \rc62.9 & \rc\textbf{88.0} & \rc54.6 & \rc57.6 & \rc\textbf{58.6}\\ 
        \midrule
        \multirow{1}{3em}{\rotatebox[origin=c]{0}{LAION-400M}}  
        &  OpenCLIP$^{\dag}$~\cite{cherti2023reproducible} & 86.1 & 91.7 & 71.4 & 69.0 & 83.7 & 17.9 & 50.7 & 89.3 & 90.1 & 69.3 & 67.1 & 71.5 \\
        \midrule
        \multirow{1}{3em}{\rotatebox[origin=c]{0}{{DataComp-1B}}}  
        &  OpenCLIP$^{\dag}$~\cite{cherti2023reproducible} & 90.5 & 96.3 & 82.2 & 70.0 & 88.9 & 29.8 & 57.7 & 92.8 & 91.8 & 75.7 & 73.5 & 77.2 \\
        \midrule
        \multirow{1}{3em}{\rotatebox[origin=c]{0}{{LAION-2B}}}  
        &  OpenCLIP$^{\dag}$~\cite{cherti2023reproducible} & 86.6 & 94.9 & 76.9 & 70.5 & 88.5 & 26.8 & 56.5 & 90.2 & 90.9 & 71.4 & 70.2 & 74.9 \\\midrule
        \multirow{2}{3em}{\rotatebox[origin=c]{0}{{2.5B}}}  
        &  MetaCLIP$^{*}$~\cite{xu2023demystifying} & 88.3 & 95.7 & 79.0 & 68.5 & 82.9 & 30.3 & 62.1 & 91.7 & 93.3 & 73.9 & 72.1 & 76.2 \\
        &  Llip$^{*}$~\cite{lavoie2024modeling} & 89.0 & 95.7 & 81.4 & 70.9 & 88.2 & 41.5 & 63.7 & 93.5 & 94.7 & 74.9 & 75.3 & 79.0 \\
        \bottomrule[1.2pt]
    \end{tabular}}
    \caption{\textbf{Zero-shot classification results} in terms of top-1 accuracy on ImageNet~\cite{deng2009imagenet} and Food101~\cite{bossard2014food}, CIFAR-10~\cite{krizhevsky2009learning}, CIFAR-100~\cite{krizhevsky2009learning}, SUN397~\cite{xiao2010sun}, Stanford Cars~\cite{KrauseStarkDengFei-Fei_3DRR2013}, FGVCAircraft~\cite{maji2013fine}, DTD~\cite{cimpoi2014describing}, Oxford Pets~\cite{parkhi2012cats}, Caltech101~\cite{fei2004learning}, Flowers102~\cite{nilsback2008automated}, and ImageNet~\cite{deng2009imagenet}. The vision encoder architecture is ViT-B/16. The best results are highlighted in \textbf{bold}. Results are reproduced with our setup for fair comparison unless otherwise marked. $^{*}$: Results copied from their work. $^{\dag}$: Results obtained using their official pre-trained weights.}
    \label{tab:zs_cla}
    \vspace{-3mm}
\end{table*}

\noindent
\textbf{Pre-training Datasets.} We pre-train \method on Conceptual Caption 3M (CC3M) and 12M (CC12M)~\cite{sharma2018conceptual} as well as YFCC15M~\cite{cui2022democratizing}, a subset of YFCC100M~\cite{thomee2016yfcc100m}. The combination of these three datasets (CC3M, CC12M, and YFCC15M) results in a merged dataset of 30M pairs, referred to as Merged-30M, following DreamLIP~\cite{zheng2025dreamlip}. As mentioned in~\cref{sub:aug}, we utilize synthetic long captions to obtain multi-modal augmentation. For CC3M, CC12M and YFCC15M, we use the synthetic captions generated by DreamLIP~\cite{zheng2025dreamlip}. The new captions were obtained using InstructBLIP~\cite{instructblip}, LLaVA-1.5~\cite{liu2024improved}, and ShareGPT4V~\cite{chen2023sharegpt4v}.

\noindent
\textbf{Implementation Details.} COSMOS is built on the OpenCLIP~\cite{cherti2023reproducible} code repository, where we primarily adopt the default settings of their implementation. We use ViT-B/16 as the vision encoder designed for $224\times224$ image sizes, while our CLIP text encoder has the sequence length of 77. Each encoder consists of 12 transformer layers, and the embedding size is fixed at 512. We train \method on A100 GPUs for 32 epochs, using a learning rate of $5 \times 10^{-4}$ and a batch size of 1,024 for CC3M, or 4,096 for CC12M, YFCC15M and Merged-30M. Following DINO~\cite{caron2021emerging}, we crop two global image views at scales between 0.4 and 1.0, resizing them to $224\times224$ pixels, and six local image views at scales between 0.05 and 0.4, resized to $96\times96$ pixels. Similarly, the two global text views consist of one to five sentences randomly sampled from the synthetic captions, while the six local text views consist of a single sentence sampled from the same caption set. The teacher momentum $\lambda$ is empirically set to 0.999 for CC3M and 0.99 for other datasets (more details in \cref{subsec:supp_hyper}). 

\subsection{Zero-Shot Image-Text Retrieval}

As shown in \cref{tab:zs_ret}, \method achieves 53.1\% and 40.1\%  R@1 score for image-to-text (I2T) and text-to-image (T2I) retrieval on MSCOCO~\cite{lin2014microsoft}. These results are 12.3\% and 10.3\% higher than DreamLIP~\cite{zheng2025dreamlip} (40.8\% and 29.8\%, respectively). When we switch to a larger pre-training set (CC12M), \method further improves its performance, reaching 64.2\% and 48.9\% R@1 scores, outperforming  CLIP~\cite{radford2021learning} by 8.0\% and 9.1\%, respectively. When we employ YFCC15M~\cite{cui2022democratizing} as the pre-training dataset, \method attains a performance of 67.5\% and 50.9\% in terms of I2T and T2I R@1 scores. The highest performance of 68.0\% and 52.5\% on the MSCOCO~\cite{lin2014microsoft} dataset is obtained by \method, when Merged-30M is used for pre-training. \method also achieves the best performance for image-text retrieval on the Flickr30k~\cite{young2014image} dataset, achieving an R@1 score of 92.9\% and 80.3\% in terms of I2T and T2I retrieval when Merged-30M is used for pre-training, while SigLIP~\cite{zhai2023sigmoid} attains only  89.9\% and 73.3\%, respectively. Overall, \method outperforms all baselines on both datasets (Flickr30k~\cite{young2014image} and MSCOCO~\cite{lin2014microsoft}) in the image-text retrieval task. 

We include other baselines for state-of-the-art comparison, including CLIP-based models (e.g., OpenCLIP~\cite{cherti2023reproducible}, MetaCLIP~\cite{xu2023demystifying}, and Llip~\cite{lavoie2024modeling}) trained on larger datasets and models based on text augmentation (e.g., LaCLIP~\cite{fan2024improving} and MLLM-A~\cite{liu2023mllms}) and self-supervised learning (e.g., SLIP~\cite{mu2022slip}, MaskCLIP~\cite{dong2023maskclip}, and SF-CLIP~\cite{sameni2024building}). It is noteworthy that \method trained on Merged-30M even outperforms Llip~\cite{fan2024improving} trained on 2.5B image-text pairs (68.0\% vs 63.4\% and 52.5\% vs 45.6\% in terms of R@1 I2T and T2I retrieval on MSCOCO). \method's improvement stems from the cross-modality distillation, since the distillation signal is passed through both encoders encouraging multi-modal alignment.
 
\begin{table*}[t]
    \centering
    \resizebox{\linewidth}{!}{
    \begin{tabular}{lcccccccccc}
    \toprule
    \multirow{2}{*}{\textbf{Method}} & \multirow{2}{*}{\textbf{Data Size}} & \multicolumn{3}{c}{\it With background category} & \multicolumn{5}{c}{\it Without background category} & \multirow{2}{*}{Avg.} \\\cmidrule(lr){3-5}\cmidrule(lr){6-10}
    & & VOC21 & Context60 & COCO-Obj. & VOC20 & City. & Context59 & ADE & COCO-Stf. \\\midrule
    OpenCLIP$^{\dag}$~\cite{cherti2023reproducible} & 400M &  18.4    &  9.1   &  8.7   &  43.8   &  6.9   &  9.8   &  2.8   &  6.9    &  13.3 \\
    OpenCLIP$^{\dag}$~\cite{cherti2023reproducible} & 1B &  22.5    &  11.9   &  11.4   &  51.3   &  8.5   &  12.6   &  4.2   &  9.2    &  16.5 \\
    OpenCLIP$^{\dag}$~\cite{cherti2023reproducible} & 2B &  19.8    &  9.6   &  9.9   &  51.3   &  6.2   &  10.2   &  3.5   &  7.2    &  14.7 \\
    \rc\method         &\rc30M     & \rc\textbf{29.6}    & \rc\textbf{12.5}   & \rc\textbf{15.2}   & \rc\textbf{53.6}   & \rc\textbf{13.9}    & \rc\textbf{15.7}   & \rc\textbf{8.5}    & \rc\textbf{10.7}    & \rc\textbf{20.0}  \\\midrule
    CLIP$^{*}$~\cite{radford2021learning} &400M& 18.8 & 9.9 & 8.1 & 49.4 & 6.5 & 11.1 & 3.1 & 5.7 & 14.1 \\
    MaskCLIP$^{*}$~\cite{zhou2022extract} &400M& 43.4 & 23.2 & 20.6 & 74.9 & 24.9 & 26.4 & 11.9 & 16.7 & 30.3 \\
    SCLIP$^{*}$~\cite{wang2023sclip}      &400M& \textbf{59.1} & 30.4 & 30.5 & \textbf{80.4} & 32.2 & \textbf{34.2} & 16.1 &  22.4 & \textbf{38.2}\\
    SILC-C$^{*}$\cite{naeem2023silc}      &10B&  -   &  -   &  -   & 77.5 & 26.9 & 31.6 & \textbf{19.3} & 20.8  &  -    \\
    \rc\method w/ SCLIP &\rc30M      & \rc52.8    & \rc\textbf{31.2}    & \rc\textbf{31.3}     & \rc77.7    & \rc\textbf{34.7}  & \rc33.7    & \rc17.7   &  \rc\textbf{23.2}   & \rc37.8     \\
    \bottomrule
    \end{tabular}}
    \caption{\textbf{Zero-shot semantic segmentation results} in terms of mIoU (mean Intersection over Union) on the PASCAL VOC 2012~\cite{everingham2015pascal}, PASCAL Context~\cite{mottaghi2014role}, Cityscapes~\cite{cordts2016cityscapes}, ADE20k~\cite{zhou2019semantic}, COCO-Stuff, and COCO-Object~\cite{caesar2018coco} datasets with and without background category. The vision encoder architecture is ViT-B/16. We also report the performance of our method enhanced with the SCLIP mechanism~\cite{wang2023sclip}.
    The best results are highlighted in \textbf{bold}. $^{*}$: Results copied from their work. $^{\dag}$: Results obtained using their official pre-trained weights.}
    \label{tab:zs_seg}
    \vspace{-3mm}
\end{table*}

\subsection{Zero-Shot Image Classification}

We observe in~\cref{tab:zs_cla} that \method outperforms the baselines CLIP~\cite{radford2021learning}, SigLIP~\cite{zhai2023sigmoid}, DreamLIP~\cite{zheng2025dreamlip},  MaskCLIP~\cite{dong2023maskclip}, and SF-CLIP~\cite{sameni2024building} on most of the evaluation datasets with the same pre-training dataset. For instance, \method pre-trained on CC3M reaches much higher accuracy of 37.1\% on ImageNet compared to CLIP (23.9\%) and SigLIP (25.9\%). As expected, the classification task greatly benefits from the pre-training dataset size, i.e., \method pre-trained on CC12M achieves 51.4\% on ImageNet while on Merged-30M, it reaches 57.6\%. 

Additionally, \method surpasses all baselines regardless of the pre-training set, in terms of average accuracy computed over the 11 datasets. For instance, when pre-trained on YFCC15M, \method achieves an average accuracy of 50.0\%, while SigLIP~\cite{zhai2023sigmoid} and DreamLIP~\cite{zheng2025dreamlip} reach 44.2\% and 46.3\%, respectively. It is worth noting that on two fine-grained datasets, i.e., Cars and SUN397, the accuracy improvements of \method trained on Merged-30M is significant (67.8\% vs 64.7\% on SUN397, 31.2\% vs 23.4\% on Cars compared to DreamLIP).

Models trained on billions of samples, e.g., OpenCLIP~\cite{cherti2023reproducible}, MetaCLIP~\cite{xu2023demystifying}, and Llip~\cite{lavoie2024modeling}  achieve higher accuracy, implying that the larger the number of images the higher the accuracy for the classification task, since the model is required to learn the global information.

\subsection{Zero-Shot Semantic Segmentation}

As shown in~\cref{tab:zs_seg} top, \method trained on only 30M samples outperforms the best OpenCLIP model~\cite{cherti2023reproducible} trained on 1B samples in all benchmarks. In the datasets with a background category, \method (29.6\%, 12.5\%, and 15.2\%, respectively)  surpasses the performance of OpenCLIP~(1B) model (22.5\%, 11.9\%, and 11.4\%, respectively) by a large margin. Remarkably, \method almost doubles the performance on the fine-grained Cityscapes dataset~\cite{cordts2016cityscapes} and achieves 20.0\% average mIoU over all segmentation benchmarks compared to 16.5\% obtained by the OpenCLIP model~(1B). While CLIP-based models often suffer from feature suppression, \method alleviates this issue through cross-modality embeddings and local-to-global matching.

In addition, we apply SCLIP~\cite{wang2023sclip} to our method where the weight of the last transformer block is adapted to construct the Correlative Self-Attention (CSA) block. \citet{wang2023sclip} claimed that the CSA block maximizes the localization of patch embeddings without extra training. Although the CSA mechanism of SCLIP is tailored for the original CLIP model, it also improves the segmentation results of \method. As a result, our \method, trained on 30M samples, achieves an average mIoU of 37.8\%, which is comparable to the 38.2\% achieved by SCLIP~\cite{wang2023sclip}, which was trained on 400M samples. Notably, \method outperforms SILC~\cite{naeem2023silc} in several benchmarks despite a significantly smaller training budget (30M vs 10B samples). For instance, our model (vs SILC) achieves 34.7\% and 23.2\% (vs 26.9\% and 20.8\%) on the Cityscapes and COCO-stuff datasets. These results suggest that \method can effectively generate fine-grained multi-modal representations with a much smaller dataset size. In \cref{subsec:supp_add_seg}, we present additional segmentation results comparing COSMOS to baselines trained on Merged-30M.

\begin{table}[t]
    \centering
    \setlength\tabcolsep{3pt}
    \scalebox{0.78}{
    \begin{tabular}{lc|ccc|c|ccc|c}
    \toprule
    \multirow{2}{*}{Method} & Data & \multicolumn{4}{c|}{SugarCrepe} & \multicolumn{3}{c|}{SVO}  & \multicolumn{1}{c}{MMVP} \\
             &  Size & Re. & Sw. & \multicolumn{1}{c}{Add} & Avg & Sub. & Verb & Obj. & Avg\\\midrule
    CLIP~\cite{radford2021learning} & 30M & 76.2 & 68.5  & 79.5  & 76.5 & 52.2 & 36.4   & 62.0    & 18.5 \\ 
    SigLIP~\cite{zhai2023sigmoid}   & 30M & 76.6 & 67.8  & 77.9  & 76.0 & 52.3 & 36.3   & 61.5   & 20.7 \\
    DreamLIP~\cite{zheng2025dreamlip}     & 30M & 82.7 & 76.5  & 82.1 & 81.8 & 57.1 & 42.2  & 66.5   & 25.2\\
    \rc\method                             & \rc30M & \rc\textbf{87.3} & \rc\textbf{79.1}  & \rc\textbf{88.1}  & \rc\textbf{86.6} & \rc\textbf{59.0}  & \rc\textbf{45.1}   & \rc\textbf{68.5}   & \rc\textbf{25.9}\\
    \midrule
            & 400M & 81.6 & 65.2  & 82.5 & 80.0 & 58.5 & 37.1  & 66.0 &   \textbf{25.9}\\
    OpenCLIP~\cite{cherti2023reproducible} & 1B & 82.7 & 64.6  & 86.6 & 81.9 & 55.5 & 35.3  & 62.9  & \textbf{25.9}\\
      & 2B & 84.7 & 67.0  & 87.4 & 83.5 & 58.5 & 39.5  & 65.3 & 21.5\\
    \bottomrule
    \end{tabular}}
    \caption{\textbf{Robustness assessment.} We evaluate \method on SugarCrepe~\cite{hsieh2024sugarcrepe} (Replace = Re., Swap = Sw., Add), SVO~\cite{hendricks2021probing} (Subject = Sub., Verb, Object = Obj.), and MMVP-VLM~\cite{tong2024eyes}.}
    \label{tab:visual_perception}
    \vspace{-3mm}
\end{table}

\subsection{Visual Perception \& Contextual Understanding}

We evaluate \method on the visual perception and contextual understanding tasks~\cite{tong2024eyes, hsieh2024sugarcrepe, hendricks2021probing} to demonstrate the robustness of our representations. \method and previous methods~\cite{radford2021learning,zhai2023sigmoid,zheng2025dreamlip} trained on Merged-30M datasets are compared in \cref{tab:visual_perception}, alongside the OpenCLIP~\cite{cherti2023reproducible} models trained on larger datasets.
These tasks are specifically designed to assess the perception capabilities of Vision-Language Models, as both visual and text input must be understood in detail to solve them.

Increasing the data size generally improves the performance of OpenCLIP models~\cite{cherti2023reproducible} as shown by the SugarCrepe results. Interestingly, \method not only surpasses the previous baselines, but also outperforms the OpenCLIP models, demonstrating that self-distillation and cross-modality learning play crucial roles in the compositional understanding of multi-modal data. It is also noteworthy that \method (trained on 30M samples) achieves the same performance (25.9\%) as OpenCLIP~\cite{cherti2023reproducible} (trained on 1B samples) on the challenging benchmark MMVP-VLM~\cite{tong2024eyes}. This result suggests that \method utilizes the pre-training data more efficiently, being able to learn distinguishing features with less data. Additionally, on the SugarCrepe dataset~\cite{hsieh2024sugarcrepe}, where the task is to choose the correct caption among two captions (e.g., in the wrong caption some words are replaced, swapped inside the sentence or added), \method achieves an average accuracy of 86.6\%. This result signifies that \method is able to differentiate between subtle differences in the text and, to a great extent, it does not treat the text as a bag-of-words. In \cref{subsec:supp_mllm}, we presents results in the MLLM setting by integrating our vision encoder in the LLaVA framework~\cite{liu2024improved}.

\subsection{Ablation Study}
\begin{table}[t]
    \centering
    \setlength\tabcolsep{2.5pt}
    \scalebox{0.85}{
    \begin{tabular}{llcccccccc}
    \toprule
    &\multirow{2}{*}{\textbf{Method}} & \multicolumn{1}{c}{ImageNet} & \multicolumn{2}{c}{MSCOCO} & \multicolumn{2}{c}{Flickr30K} \\\cmidrule(lr){3-3}\cmidrule(lr){4-5}\cmidrule(lr){6-7}
                      &          & Top-1 & I2T & T2I & I2T & T2I \\\midrule
    \rownumber{1}&CLIP                        & 17.5 & 15.0  & 10.7 & 31.8  &21.3 \\
    \rownumber{2}&  { } + Image Aug.  & 19.1  &  17.5 & 12.7  &36.4 & 27.6\\
    \rownumber{3}&{ } + EMA                 & 19.1  & 17.5 & 12.7  &36.5 & 27.6\\ 
    \rownumber{4}&{ } + Image Self-distill. & 22.8  & 20.6 & 14.6  & 40.2 & 29.4\\
    \rownumber{5}&{ } + Text Aug.      & 34.4  &  50.4 & 37.5 &  81.3  & 65.5 \\
    \rownumber{6}&{ } + Text Self-distill.  & 35.4  &  51.0 & 38.0  & 82.5  & 67.1 \\
    \rc\rownumber{7}& \rc{ } + Cross-attn. (=\method) & \rc\textbf{37.1} & \rc\textbf{53.1}  & \rc\textbf{40.1}  & \rc\textbf{84.1}  & \rc\textbf{68.6} \\ 
    \bottomrule
    \end{tabular}
    }
    \caption{\textbf{Component ablation.} We show top-1 accuracy on ImageNet~\cite{deng2009imagenet}, and  R@1 score for I2T and T2I retrieval tasks on  Flickr30K~\cite{young2014image} and MSCOCO~\cite{lin2014microsoft}.
    All models are trained on CC3M with the batch size of 1,024 and ViT-B/16 image encoder.} 
    \label{tab:ablation1}
    \vspace{-3mm}
\end{table}

We validate the effectiveness of each component in \cref{tab:ablation1}, starting with a naive CLIP model trained on image and text pairs without augmentation (row~\rownumber{1}).
This simple setup achieves an accuracy of 15.0\% and 10.7\% in terms of zero-shot performance on MSCOCO image-to-text and text-to-image tasks. Applying image augmentation (row~\rownumber{2}) improves all metrics, while EMA (row~\rownumber{3}) shows marginal effects on the results.  Applying self-distillation on the image encoder brings additional improvements, as it enforces local-to-global correspondence (row~\rownumber{4}).
Row~\rownumber{5} reveals that employing long synthetic captions as text augmentation plays an important role, boosting the accuracy on MSCOCO to 50.4\% and 37.5\%. Simply adapting self-distillation for the text encoder (row~\rownumber{6}) slightly improves the performance (51.0\% and 38.0\% on MSCOCO). The self-distillation mechanism is further enhanced by applying a cross-attention module (row~\rownumber{7}) reaching the best accuracy of 53.1\% and 40.1\% on MSCOCO retrieval task.

Additionally, we reproduce previous self-supervised methods based on CLIP, such as SILC~\cite{naeem2023silc} and SLIP~\cite{mu2022slip}, using the same synthetic dataset on which \method is trained. We directly compare these methods with \method reporting the results in \cref{subsec:supp_comparison}.

\subsection{Qualitative Results}

\begin{figure}[t]
    \centering
    \includegraphics[width=1.0\linewidth]{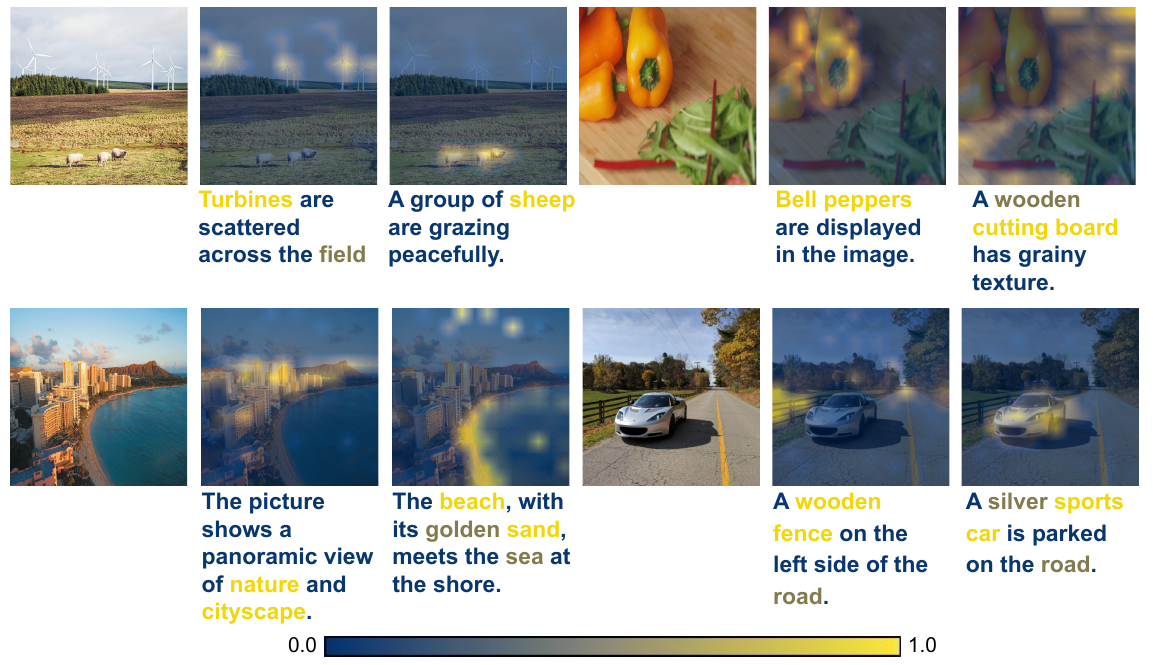}
    \caption{Visualization of attention map in cross-attention modules. Attention weights are normalized between 0 and 1.} 
    \label{fig:qualitative}
    \vspace{-3mm}
\end{figure}

We present additional qualitative results of \method in \cref{fig:qualitative}. We visualize the normalized attention weights of the cross-attention modules for both images and text. For clarity, we use only three colors for the captions, discretized by their original weight values (dark blue for low value, light green for intermediate value, and yellow for high value). Image attention results are based on the input text, while text attention results are conditioned on the input image. 

The qualitative results illustrate that various objects are highlighted in both the image and the text, regardless of their location and size. In the top-left example, the sheep from the foreground and turbines from the background are accurately captured in both image and caption, despite occupying only a small part of the image. Additionally, our module can attend to specific attributes of objects, such as the wooden fence or the silver sports car in the bottom-right example. More qualitative results are in \cref{sec:supp_qualitative}.

\section{Conclusion}

Our \method, a model for vision-language pre-training, uses cross-modality representation for self-distillation. Specifically, we design a text-cropping strategy based on long caption datasets, which enable us to learn local-to-global text correspondences. Our cross-attention module for self-supervised learning further encourages our model to learn both visual and textual grounding, overcoming the previous shortcomings of the contrastive vision-language pre-training. Experimental results on various downstream tasks demonstrate that \method outperforms strong baselines trained on much larger datasets on image-text retrieval, semantic segmentation and compositional understanding.

{
\noindent
\textbf{Acknowledgements.} This work was partially funded by the ERC (853489 - DEXIM) and the Alfried Krupp von Bohlen und Halbach Foundation, which we thank for their generous support. The authors gratefully acknowledge the Gauss Centre for Supercomputing e.V. (www.gauss-centre.eu) for funding this project by providing computing time on the GCS Supercomputer JUWELS~\cite{JUWELS} at Jülich Supercomputing Centre (JSC). We would also like to thank Shyamgopal Karthik and the entire EML group for helpful feedback and discussions.
}

{
    \small
    \bibliographystyle{ieeenat_fullname}
    \bibliography{main}
}

\clearpage
\setcounter{page}{1}
\setcounter{section}{0}

\renewcommand{\thesection}{\Alph{section}}

\twocolumn[{
\renewcommand\twocolumn[1][]{#1}
\maketitlesupplementary
\begin{center}
  \centering
  \captionsetup{type=figure}
  \includegraphics[width=\linewidth]{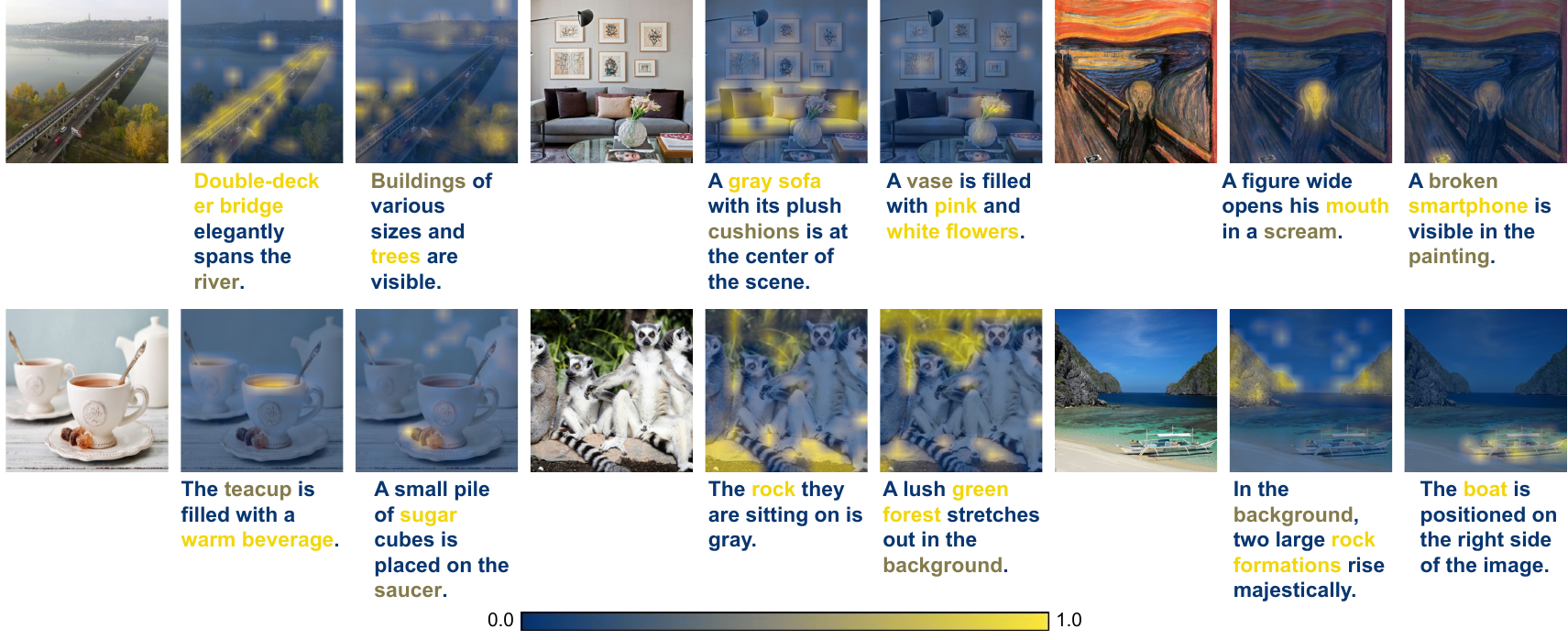}
  \captionof{figure}{\textbf{Visualization of Attention Map.} For different set of captions, we visualize the attention weights of the image and text cross-attention modules. The patch-wise (image) and token-wise (caption) attention weights are both normalized between 0 and 1.}
  \label{fig:sup_qualitative}
\end{center}
}]

In this supplementary material, we include detailed information about the training procedure of \method. We first provide additional qualitative results in~\cref{sec:supp_qualitative}. The dataset configurations are described in~\cref{sec:supp_data}, and the experimental setups are outlined in~\cref{sec:supp_exp}. Additionally, we elaborate on the baseline configurations in~\cref{sec:supp_baseline} and present further experiments in~\cref{sec:supp_extra}.

\section{Qualitative Results}~\label{sec:supp_qualitative}

In \cref{fig:sup_qualitative}, we visualize the attention maps of the cross-attention modules with different captions to illustrate their focus areas. We first normalize the attention weights across all patches and map them back onto the original image. Similarly, tokens in the captions are colored based on the normalized attention weights. The qualitative results show that our cross-attention module enhances the learning of local representations in both modalities, namely image and text. For example, our model is capable of detecting relatively small objects in the image, such as sugar cubes (first example of the second row) or a broken smartphone (third example of the first row). This information is often overlooked by the previous models due to feature suppression.

\section{Dataset Configuration}\label{sec:supp_data}
As \method is trained on web datasets, we had to download the images based on the corresponding URLs, but some of the URLs were invalid. Consequently, there is a discrepancy between the number of samples used to train our models and the original dataset sizes. The specific number of samples are reported in \cref{tab:sample_num}. We lost approximately 8.6\% (YFCC15M) to 19.4\% (CC12M) of samples compared to the original datasets due to expired URLs. This detail should be considered when directly comparing our models with others.

\begin{table}[h]
    \centering
    \resizebox{\linewidth}{!}{\begin{tabular}{lcccccc}
    \toprule
          Dataset  & Original & Ours & Difference & Percentage \\\midrule
    CC3M~\cite{sharma2018conceptual}                & $3,\!318,\!333$ & $2,\!823,\!019$  & $500,\!229$  & $85.1\%$ \\
    CC12M~\cite{changpinyo2021conceptual}         & $12,\!423,\!374$ & $10,\!010,\!225$  & $2,\!413,\!149$ & $80.6\%$ \\
    YFCC15M~\cite{cui2022democratizing}           & $15,\!388,\!848$ & $14,\!065,\!827$  & $1,\!323,\!021$ & $91.4\%$\\
    Merged-30M                                    & $31,\!130,\!555$ & $26,\!899,\!071$  & $4,\!236,\!399$ & $86.4\%$\\
    PixelProse~\cite{singla2024pixels}           & $16,\!896,\!423$ & $15,\!037,\!386$  & $1,\!859,\!037$ & $89.0\%$\\
    \bottomrule
    \end{tabular}}
    \caption{\textbf{Size of the pre-training datasets.} We compare the original dataset sizes and the actual number of samples used to train our models. We also report the percentage of samples that we succeeded to retrieve from the original image URL links.}
    \label{tab:sample_num}
\end{table}

Additionally, \method is evaluated on SugarCrepe~\cite{hsieh2024sugarcrepe}, SVO~\cite{hendricks2021probing}, and MMVP~\cite{hendricks2021probing}, to assess the robustness of its multi-modal representations. 

The \textbf{SugarCrepe} dataset~\cite{hsieh2024sugarcrepe} consists of images with positive and negative captions, where the model is required to choose the positive one given an input image (50\% is the random chance). Negative captions are slightly different from the positive captions in terms of attributes and objects, with the goal of confusing the model. Since SugarCrepe is based on the COCO~\cite{lin2014microsoft} validation set, we were able to download all images without any data loss.

\textbf{SVO}~\cite{hendricks2021probing} consists of 36,841 data pairs, where each caption is paired with two images (one positive and one negative). Each image has its own triplet (subject, verb, object), and the positive and negative images differ in one component of the triplets. The original task is to match the caption with one of the two images based on the similarity scores given by VLMs (50\% by random chance). Since naive CLIP~\cite{radford2021learning} already achieves 80\% accuracy, we increased the difficulty of this task by constructing negative captions. Based on the triplet of positive and negative images, we replaced the subject, verb, or object in the positive caption with the negative one. Currently, the model not only has to match the positive image with the positive caption but also match the negative image with the negative caption (25\% by random chance). As SVO is released with image URL links, we were able to collect 22,220 data pairs (60.3\% of the original size).

\textbf{MMVP-VLM}~\cite{hendricks2021probing} contains 15 image-text pairs across 9 categories, resulting in a total of 135 image-text pairs. It categorizes 9 challenging visual patterns that most VLMs struggle with. Each data point consists of two images and two captions, requiring the VLMs to match the correct image and caption respectively (25\% by random chance). Since \citet{hendricks2021probing} released their dataset including images and captions, we were able to fully download the MMVP-VLM dataset as originally intended.

\section{Experiment Configuration}\label{sec:supp_exp}
In this section, we detail the configuration of our experiments, including hyperparameters for training (\cref{subsec:supp_hyper}), the inference process for each downstream task (\cref{subsec:supp_inference}), and the pseudocode of our training objective (\cref{subsec:supp_pseudocode}).

\subsection{Hyperparameters}\label{subsec:supp_hyper}

As shown in \cref{tab:hyper}, we primarily follow the hyperparameters described in DreamLIP~\cite{zheng2025dreamlip}, with the exception of batch size. DreamLIP utilizes a batch size of 1,024 for CC3M and 8,192 for other datasets. Due to computational constraints, we adopt a batch size of 1,024 for CC3M and 4,096 for other datasets to train \method. For reproducing CLIP~\cite{radford2021learning} and SigLIP~\cite{zhai2023sigmoid} with our setting, we use a batch size of 1,024 for CC3M and 6,144 for other datasets to establish a strong baseline.

\begin{table}[h]
    \centering
    \resizebox{0.8\linewidth}{!}{\begin{tabular}{l|l}
    \toprule
    Config & Value \\\midrule
    Optimizer & AdamW~\cite{loshchilov2017decoupled} \\
    Learning rate & $5\times10^{-4}$ \\
    Weight decay & $0.5$ \\
    Adam $\beta$ & $\beta_1, \beta_2=(0.9, 0.98)$\\
    Adam $\epsilon$ & $1\times10^{-8}$ \\
    Total epochs & $32$ \\
    Warm up iterations & $2,000$\\
    Learning rate schedule & cosine decay \\
    \bottomrule
    \end{tabular}}
    \caption{\textbf{Hyperparameter configuration.}}
    \label{tab:hyper}
\end{table}

As mentioned in the main paper, the teacher model is updated at each iteration using the exponential moving average (EMA) of the student model. To obtain an effective teacher for self-distillation, we need to determine the momentum parameter $\lambda$, which controls the update rule of the teacher parameter $\theta_t$ based on the student parameter $\theta_s$ (i.e., $\theta_t = \lambda\theta_t + (1-\lambda)\theta_s$). According to \citet{caron2021emerging}, a higher batch size requires a lower momentum parameter. Based on their configuration, we choose 0.999 for CC3M and 0.99 for other datasets by default. We did not explicitly perform a grid search for this parameter. Empirically, we found that fixing the momentum parameter obtains a better performance, unlike \citet{caron2021emerging}, who used a cosine scheduler for the momentum parameter which eventually converges to 1.0.

\subsection{Inference}\label{subsec:supp_inference}

As \method is applied to various downstream tasks, it is crucial to establish a consistent evaluation protocol for each task. Notably, the cross-attention module is not involved in the inference process. Therefore, we are able to evaluate \method as a standard CLIP-based model, utilizing the class token $[\text{cls}]$ and end-of-text token $[\text{eot}]$ as image and text embeddings, respectively.

For \textbf{zero-shot classification}, we follow the process established by CLIP~\cite{radford2021learning}. First, we construct prompts using the class label names for each dataset, as referenced in ALIP~\cite{yang2023alip}. For each class, text embeddings are generated by the text encoder using different prompts, which are then averaged to obtain the final text embedding. Given an input image, the image encoder extracts the image embedding and calculates the cosine similarity scores between the image embedding and the text embeddings of all predefined classes in each dataset. The class label with the highest score is selected as the prediction.

\textbf{Zero-shot retrieval} is based on the image and text embeddings extracted from the pre-trained models, following the evaluation scheme presented in~\cite{yang2023alip}. For each image-text pair, the pre-trained model generates image and text embeddings, collecting all embeddings for the entire dataset. We then compute a full cosine similarity matrix between all image embeddings and all text embeddings. Captions with the highest similarity scores are selected for each image to calculate the R@1, R@5, and R@10 metrics for image-to-text retrieval. Similarly, for text-to-image retrieval, we choose images with the highest scores for each caption. As each image in the validation or test set is equipped with multiple captions, image-to-text retrieval scores are generally higher than text-to-image scores.

In \textbf{zero-shot semantic segmentation}, we exclude the background category for PASCAL VOC~\cite{everingham2015pascal} and PASCAL Context~\cite{mottaghi2014role}, following \citet{cha2023learning} and \citet{wang2023sclip}. To be specific, we denote the original datasets with a background class as VOC21 and Context60, while the variants without the background category are referred to as VOC20 and Context59 in the main table. At inference time, for a given set of classes in the datasets, we obtain the corresponding text embeddings by querying our text encoder with a standard prompt. We compute the cosine similarity between the image patch embeddings (image tokens) and the text features of each class name to generate a segmentation map in a zero-shot manner. We adhere to the evaluation protocol established by SCLIP~\cite{wang2023sclip}, including specifications for the window size, stride, and other parameters. We believe that the raw segmentation output of a VLM accurately reflects its zero-shot performance; therefore, we do not fine-tune our model or apply any post-refinement techniques such as PAMR~\cite{araslanov2020single}.

To \textbf{evaluate on SugarCrepe~\cite{hsieh2024sugarcrepe}, SVO~\cite{hendricks2021probing}, and MMVP-VLM~\cite{tong2024eyes}}, we primarily referred to their evaluation demo prompts. Image and text embeddings are extracted from the pre-trained models, and then image-text pairs with higher cosine similarity scores are chosen as the final decision.

\subsection{Pseudocode}\label{subsec:supp_pseudocode}
To increase the clarity of our method, we present the pseudocode of the training objective in \cref{alg:code}. We empirically found that incorporating local image crops on the CLIP loss diminishes zero-shot performance, whereas using local text crops enhances it. Consequently, we compute the CLIP loss between all text crops and global image crops. We infer that integrating diverse captions during training allows the model to learn various objects shown in the images, while including local image crops during training likely leads to the misalignment of image-text pairs.

\begin{algorithm}[ht]
\caption{\method: Pseudocode of our loss function}
\label{alg:code}
\algcomment{
\textbf{Notes}: We assume one global view and one local view for simplicity. \texttt{sym\_nce} represents the symmetric InfoNCE loss~\cite{oord2018representation}. \texttt{cross-attn} denotes the cross attention module which requires key, value ($kv$) and query ($q$) as inputs.
}
\definecolor{codeblue}{rgb}{0.25,0.5,0.5}
\definecolor{codekw}{rgb}{0.85, 0.18, 0.50}
\begin{lstlisting}[language=python]
# img_g, img_l: Global&local crop of image
# txt_g, txt_l: Global&local crop of text
# Is, Ts: Student image&text encoder
# It, Tt: Teacher image&text encoder

# Generate embeddings of size [batch, seq_len, dim]
s_img_g, s_img_l = Is(img_g, img_l) # Student image
s_txt_g, s_txt_l = Ts(txt_g, txt_l) # Student text
t_img_g = It(img_g) # Teacher image
t_txt_g = Tt(txt_g) # Teacher text

# Split into ([CLS],img_tok) or ([EOT], txt_tok)
s_cls_g, s_img_tok_g = s_img_g
s_eot_g, s_txt_tok_g = s_txt_g
s_cls_l, _ = s_img_l
s_eot_l, _ = s_txt_l
t_cls_g, _ = t_img_g
t_eot_g, _ = t_txt_g

# Calculate CLIP loss
clip_loss = (sym_nce(s_cls_g, s_eot_g) + 
                    sym_nce(s_cls_g, s_eot_l))/2


# Generate cross-modality embeddings
s_cls = {s_cls_g, s_cls_l} # Combine student crops
s_eot = {s_eot_g, s_eot_l} # Combine student crops
h_img = s_cls + cross-attn(q=s_cls, kv=s_txt_tok_g)
h_txt = s_eot + cross-attn(q=s_eot, kv=s_img_tok_g)

# Calculate cross-modality self-distillation loss
cosmos_loss = (sym_nce(h_img, t_cls_g) + 
                    sym_nce(h_img, t_eot_g) + 
                    sym_nce(h_txt, t_cls_g) +
                    sym_nce(h_txt, t_eot_g))/4
               
final_loss = clip_loss + cosmos_loss

\end{lstlisting}
\end{algorithm}

\section{Baseline Configuration}\label{sec:supp_baseline}
In the main tables, we evaluate both the pre-trained weights from the official code repository and our reproduction. For DreamLIP~\cite{zheng2025dreamlip}, we used the official pre-trained weights with ViT/B-16 as the vision encoder. Since they did not provide weights for the ViT/B-32 vision encoder, we referenced the results from their table as shown in \cref{tab:zs_ret_vit32} and \cref{tab:zs_cla_vit32}. For OpenCLIP~\cite{cherti2023reproducible}, we utilized the models described in \cref{tab:model_name}. Specifically, they trained their CLIP models on LAION-400M~\cite{schuhmann2021laion} until 12.8 billion examples were seen, using a batch size of 33,792 for both ViT-B/16 and ViT-B/32. For DataComp-1B~\cite{gadre2024datacomp}, the models were trained until 12.8 billion examples were seen with a batch size of 90,112. Additionally, the models were trained on LAION-2B~\cite{schuhmann2022laion} with a batch size of 88,064 (ViT-B/16) or 79,104 (ViT-B/32) until 34 billion examples were seen.

\begin{table}[h]
    \centering
    \resizebox{0.9\linewidth}{!}{\begin{tabular}{l|ll}
    \toprule
    Architecture & Data & Model Name \\\midrule
                & LAION-400M & laion400m\_e32\\
    ViT-B/16    & DataComp-1B  & datacomp\_xl\_s13b\_b90k \\
                & LAION-2B   & laion2b\_s34b\_b88k \\\midrule
                & LAION-400M    & laion400m\_e32\\
    ViT-B/32    & DataComp-1B  & datacomp\_xl\_s13b\_b90k \\
                & LAION-2B   & laion2b\_s34b\_b79k \\
    \bottomrule
    \end{tabular}}
    \caption{\textbf{OpenCLIP~\cite{cherti2023reproducible} model names.}}
    \label{tab:model_name}
\end{table}

\section{Extra Experiments}\label{sec:supp_extra}

\subsection{Comparison to Previous SSL Methods}\label{subsec:supp_comparison}

 \begin{figure*}[h]
     \centering
     \includegraphics[width=0.98\linewidth]{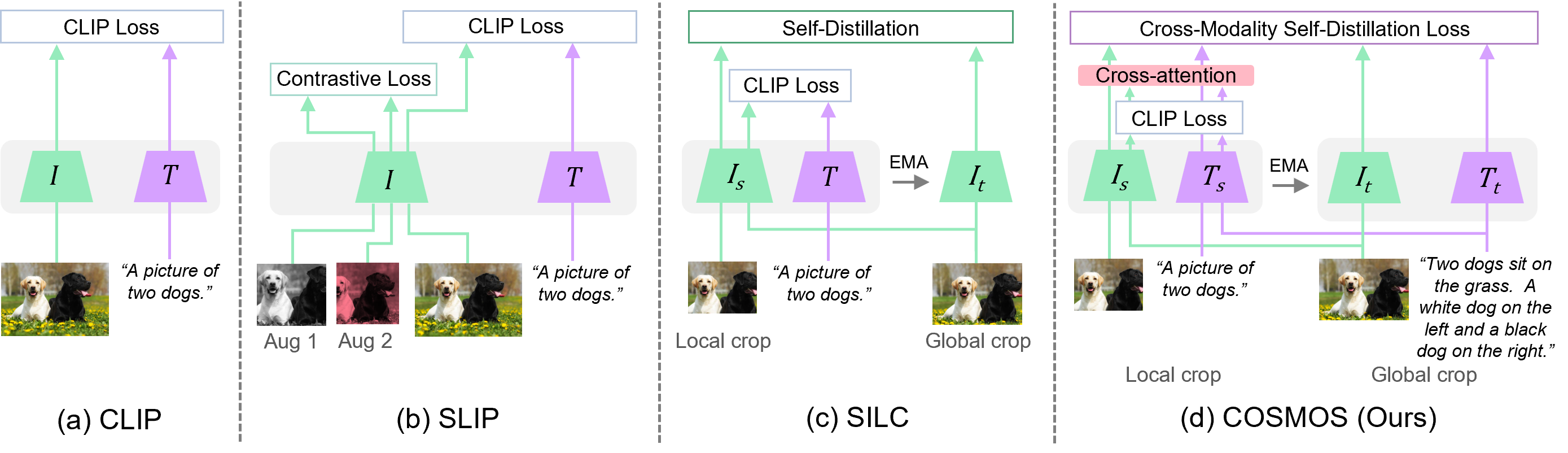}
 \caption{
     \textbf{Illustration of CLIP with self-supervised approaches.} $I$ and $T$ denote the image and text encoders, respectively. $I_t$ (or $T_t$) and $I_s$ (or $T_s$) represent the teacher and student image (or text) encoders, where the teacher is an exponential moving average (EMA) of the student. (a) CLIP~\cite{radford2021learning}: image and text embeddings are aligned during training. (b) SLIP~\cite{mu2022slip}: contrastive loss is computed on sets of two different augmentations. (c) SILC~\cite{naeem2023silc}: self-distillation loss is obtained between local and global crops of the same image. (d) \method: the cross-attention module is utilized to generate cross-modal representations which are optimized through the cross-modality self-distillation loss. We also design global and local crops of image and text pairs for effective self-supervised learning in VLMs.
 }
     \label{fig:compare}

\end{figure*}

Previous works, such as SLIP~\cite{mu2022slip} or SILC~\cite{naeem2023silc}, aim to enhance CLIP through self-supervision by explicitly optimizing local-to-global correspondences. These methods primarily focus on the image encoder. \cref{fig:compare} illustrates how our approach differs from the self-supervised contrastive language-image pre-training methods, SLIP~\cite{mu2022slip} and SILC~\cite{naeem2023silc}.

In \cref{tab:ablation3}, we directly compare \method with the previous methods depicted in \cref{fig:compare}. For a fair comparison, we re-implemented these methods based on the OpenCLIP~\cite{cherti2023reproducible} code repository and trained them on the CC3M dataset with long captions provided by DreamLIP~\cite{zheng2025dreamlip}. The projection layer parameters follow the SLIP configuration, while the optimal temperature and loss scale for SILC were carefully determined. After a grid search, we selected student and teacher temperatures of 0.1 and 0.02, respectively, and used loss scale hyperparameters of (1.9, 0.1) for CLIP loss and self-distillation loss (see \cref{tab:loss_scale_ablation}). Without extensive hyperparameter tuning, \method significantly outperforms previous methods obtaining an accuracy of 37.1\% in zero-shot classification on ImageNet~\cite{deng2009imagenet}, and 53.1\% R@1 score in image-to-text retrieval on MSCOCO~\cite{lin2014microsoft}. As mentioned in the introduction, focusing solely on enhancing image representation leads to sub-optimal results in VLMs, as shown by SILC which reached an accuracy score of only 30.4\%  on ImageNet and an R@1 score of 48.6\% on image-to-text retrieval on MSCOCO, highlighting the importance of self-distillation with cross-modality representations.

\begin{table}[t]
    \centering
    \resizebox{\linewidth}{!}{\begin{tabular}{lcccccc}
    \toprule
    \multirow{2}{*}{\textbf{Method}} & \multicolumn{1}{c}{ImageNet} & \multicolumn{2}{c}{MSCOCO} & \multicolumn{2}{c}{Flickr30k} \\\cmidrule(lr){2-2}\cmidrule(lr){3-4}\cmidrule(lr){5-6}  
                                & Top-1 & I2T@1 & T2I@1 & I2T@1 & T2I@1  \\\midrule
    (a) CLIP~\cite{radford2021learning} & 23.9 & 40.2  & 27.2  & 68.4  & 52.1    \\
    (b) SLIP~\cite{mu2022slip}          & 26.6 & 44.4  & 30.5  & 75.8  & 58.7  \\
    (c) SILC~\cite{naeem2023silc}       & 30.4 &  48.6  & 35.4 & 79.0  & 62.1  \\
    (d) \rc\method                         & \rc\textbf{37.1} & \rc\textbf{53.1}  & \rc\textbf{40.1}  & \rc\textbf{84.1}  & \rc\textbf{68.6}  \\
    \bottomrule
    \end{tabular}}
    \caption{\textbf{Comparison to methods described in \cref{fig:compare}}. All models are trained on CC3M with long synthetic caption. We use the batch size of 1,024 and ViT-B/16 image encoder.}
    \label{tab:ablation3}
\end{table}

\subsection{Experiments with ViT-B/32 Architecture}

\begin{table*}[h]
\centering
\setlength\tabcolsep{1.0pt}
\resizebox{.82\linewidth}{!}{

\begin{tabular}
{l<{\hspace{3.0em}}c<{\hspace{1.0em}} | c<{\hspace{0.7em}}c<{\hspace{0.7em}}c<{\hspace{0.7em}} | c<{\hspace{0.7em}}c<{\hspace{0.7em}}c<{\hspace{0.7em}} | c<{\hspace{0.7em}}c<{\hspace{0.7em}}c<{\hspace{0.7em}} | c<{\hspace{0.7em}}c<{\hspace{0.7em}}c<{\hspace{0.7em}}}

\toprule
&&\multicolumn{6}{c|}{Flickr30K}& \multicolumn{6}{c}{MSCOCO}\\
&&\multicolumn{3}{c|}{Image-to-text} & \multicolumn{3}{c|}{Text-to-image} & \multicolumn{3}{c|}{Image-to-text} & \multicolumn{3}{c}{Text-to-image} \\
Data & Method & R@1 & R@5 & R@10 & R@1 & R@5 & R@10 & R@1 & R@5 & R@10 & R@1 & R@5 & R@10 \\
\midrule
\multirow{4}{3em}{CC3M}
& CLIP~\cite{radford2021learning}   & 53.4 & 77.1 & 85.0 & 38.4 & 65.4 & 74.6 & 31.3 & 55.3 & 66.9 & 19.8 & 42.7 & 54.8  \\ 
& SigLIP~\cite{zhai2023sigmoid}     & 56.4 & 78.9 & 85.7 & 40.2 & 66.5 & 76.0 & 30.8 & 55.5 & 66.8 & 20.6 & 43.9 & 55.4 \\ 
& DreamLIP$^{*}$~\cite{zheng2025dreamlip} & 57.6 & 84.4 & 89.6 & 42.2 & 69.0 & 77.7 & 33.4 & 60.7 & 72.0 &  23.4 & 47.2 & 58.6 \\ 
& \rc\method                            & \rc\textbf{74.3} & \rc\textbf{92.3} & \rc\textbf{95.9} & \rc\textbf{59.2} & \rc\textbf{82.6} & \rc\textbf{89.1} & \rc\textbf{47.6} & \rc\textbf{73.1} & \rc\textbf{82.0} & \rc\textbf{33.5} & \rc\textbf{59.7} & \rc\textbf{70.6}  \\ 
\midrule
\multirow{4}{3em}{CC12M}
& CLIP~\cite{radford2021learning}      & 76.8 & 92.9 & 95.8 & 60.3 & 83.4 & 89.4 & 50.6 & 76.5 & 84.3 & 34.9 & 61.5& 72.3  \\ 
& SigLIP~\cite{zhai2023sigmoid}        & 75.8 & 92.7 & 96.1 & 60.0 & 83.1 & 89.7 & 51.2 & 76.5 & 84.7 & 35.4 & 61.9 & 72.4 \\ 
& DreamLIP$^{*}$~\cite{zheng2025dreamlip}    & 78.7 & 94.6 & 97.6 & 61.0 & 83.9 & 89.8 & 53.4 & 77.1 & 84.7 &  36.7 & 62.3 & 72.3 \\ 
& \rc\method                     & \rc\textbf{86.5} & \rc\textbf{97.5} & \rc\textbf{98.7} & \rc\textbf{69.8} & \rc\textbf{89.3} & \rc\textbf{94.1} & \rc\textbf{59.6} & \rc\textbf{82.3} & \rc\textbf{89.4} & \rc\textbf{43.0} & \rc\textbf{69.5} & \rc\textbf{78.9}  \\ 
\midrule
\multirow{4}{3em}{YFCC15M}
& CLIP~\cite{radford2021learning}      & 84.6 & 97.1 & 98.9 & 66.0 & 87.8 & 92.7 & 56.7 & 81.6 & 88.6 & 40.1 & 66.8 & 76.6  \\ 
& SigLIP~\cite{zhai2023sigmoid}        & 82.5 & 97.4 & 98.5 & 66.9 & 87.8 & 92.3 & 56.3 & 82.0 & 89.0 & 40.0 & 66.7 & 76.9 \\ 
& DreamLIP$^{*}$~\cite{zheng2025dreamlip}    & 84.9 & 97.3 & 98.7 & 66.0 & 86.4 & 91.4  & 55.7 & 80.5 & 88.2  & 39.8 & 66.0 & 75.5 \\ 
& \rc\method                            & \rc\textbf{90.2} & \rc\textbf{98.7} & \rc\textbf{99.4} & \rc\textbf{73.3} & \rc\textbf{91.6} & \rc\textbf{95.4} & \rc\textbf{64.5} & \rc\textbf{86.1} & \rc\textbf{91.8} & \rc\textbf{46.0} & \rc\textbf{72.2} & \rc\textbf{81.0}  \\ 
\midrule
\multirow{4}{3em}{\text{Merged-30M}}
& CLIP~\cite{radford2021learning}      & 85.6 & 96.7 & 99.0 & 69.5 & 89.9 & 94.3 & 59.3 & 83.1 & 89.9 & 42.8 & 69.3 & 79.0  \\ 
& SigLIP~\cite{zhai2023sigmoid}        & 88.4 & 97.7 & 99.1 & 70.7 & 90.3 & 94.6 & 59.5 & 83.3 & 90.1 & 43.8 & 70.1 & 79.6 \\ 
& DreamLIP$^{*}$~\cite{zheng2025dreamlip}    & 87.2 & 97.5 & 98.8 & 66.4 & 88.3 & 93.3 & 58.3 & 81.6 & 88.8  & 41.1 & 67.0 & 76.6  \\ 
& \rc\method                       & \rc\textbf{89.9} & \rc\textbf{98.8} & \rc\textbf{99.3} & \rc\textbf{76.1} & \rc\textbf{92.8} & \rc\textbf{96.2} & \rc\textbf{64.3} & \rc\textbf{86.5} & \rc\textbf{92.0} & \rc\textbf{48.4} & \rc\textbf{74.2} & \rc\textbf{82.6}  \\ 
\midrule
\multirow{1}{3em}{\text{LAION-400M}}
& OpenCLIP$^{\dag}$~\cite{cherti2023reproducible}  &79.7  & 95.0& 97.6 & 60.9 & 84.8 & 90.7 & 51.7 & 76.6 & 84.9 & 33.7  & 59.4& 69.9  \\
\midrule
\multirow{1}{3em}{\text{DataComp-1B}}
& OpenCLIP$^{\dag}$~\cite{cherti2023reproducible}  &80.1    &94.6   &97.2     &62.9    &85.4    &91.1   &54.6     &78.4     &85.8     &36.3    &62.1    & 72.6   \\
\midrule
\multirow{1}{3em}{\text{LAION-2B}}
& OpenCLIP$^{\dag}$~\cite{cherti2023reproducible}  &85.4 &96.2    &98.2     & 68.4    & 89.0    & 93.4   &56.6    &80.3     &87.4     &38.8    &64.8    &74.7   \\
\bottomrule          
\end{tabular}}
\vspace{-2mm}
\caption{\textbf{Zero-shot image-text retrieval results} in terms of R@1, R@5, and R@10 on the Flickr30K~\cite{young2014image} and MSCOCO~\cite{lin2014microsoft} datasets. The vision encoder architecture is ViT-B/32. The best results are highlighted in \textbf{bold}. Results are reproduced with our setup for fair comparison unless otherwise marked. $^{*}$: Results copied from their work. $^{\dag}$: Results obtained using their official pre-trained weights.}
\label{tab:zs_ret_vit32}
\end{table*}

In addition to the experiments with ViT-B/16 presented in the main paper, we conducted the same experiments using the ViT-B/32 vision encoder architecture, as shown in \cref{tab:zs_ret_vit32} and \cref{tab:zs_cla_vit32}. Overall, the improvements achieved with \method are consistent with those observed in the main table using ViT-B/16.

In the zero-shot retrieval tasks (\cref{tab:zs_ret_vit32}), \method not only surpasses previous strong baselines (CLIP~\cite{radford2021learning}, SigLIP~\cite{zhai2023sigmoid}, and DreamLIP~\cite{zheng2025dreamlip}) trained on the same datasets, but also exceeds the performance of OpenCLIP~\cite{cherti2023reproducible} models trained on much larger datasets. Notably, \method trained on Merged-30M achieves 64.3\% and 48.4\% in image-to-text and text-to-image R@1 scores on MSCOCO, significantly outperforming DreamLIP (58.3\% and 41.1\%). Furthermore, \method trained on CC12M already demonstrates higher accuracy compared to OpenCLIP trained on LAION-2B dataset. These results highlight the effectiveness of \method in generating more fine-grained and comprehensive multi-modal representations through cross-modality self-distillation.

In the zero-shot classification tasks (\cref{tab:zs_cla_vit32}), \method improves all metrics compared to CLIP~\cite{radford2021learning} and SigLIP~\cite{zhai2023sigmoid} across most datasets, while achieving comparable results to DreamLIP~\cite{zheng2025dreamlip} on YFCC15M and Merged-30M. DreamLIP~\cite{zheng2025dreamlip} aligns visual patch embeddings with positive text embeddings, promoting the learning of global structure by image tokens, which enhances the performance of classification task where the model focuses on global information. Although \method optimizes local-to-global correspondence to enhance the representation of local information through self-distillation, its performance is similar to DreamLIP. Additionally, increasing the dataset size generally improves accuracy, suggesting that the total number of images is a crucial factor for achieving high performance in classification. The results obtained with ViT-B/32 are consistent with those of ViT-B/16 from the main paper.

\subsection{Experiment on PixelProse Dataset}

To demonstrate the adaptability of \method to various captions, we employ PixelProse~\cite{singla2024pixels}, a synthetic caption dataset similar to DreamLIP. PixelProse filters and combines data from CommonPool~\cite{gadre2024datacomp}, CC12M~\cite{sharma2018conceptual}, and RedCaps~\cite{desai2021redcaps} to create over 16 million image and alt-text pairs. They used Gemini 1.0 Pro~\cite{singla2024pixels} to generate new captions. As shown in \cref{tab:pixelprose} and \cref{tab:pixelprose2}, we trained CLIP~\cite{radford2021learning}, SigLIP~\cite{zhai2023sigmoid}, and \method on PixelProse using the same setup as before. The actual number of image-text pairs used to train these models is detailed in \cref{tab:sample_num}.

In \cref{tab:pixelprose}, \method consistently outperforms the CLIP and SigLIP models by a significant margin with both ViT-B/16 and ViT-B/32 vision encoders. For instance, \method with ViT-B/16 achieves 62.4\% and 43.4\% in R@1 scores for image-to-text and text-to-image retrieval on MSCOCO. This performance even surpasses strong baselines from OpenCLIP~\cite{cherti2023reproducible}, which achieve 56.5\% and 37.9\% on LAION-400M, 58.2\% and 39.8\% on DataComp-1B, and 59.3\% and 41.7\% on LAION-2B (see Tab.\ 1 main).

Similarly, \cref{tab:pixelprose2} shows that \method outperforms previous baselines in the zero-shot classification tasks. Notably, \method trained on PixelProse achieves the best average accuracy of 60.7\% with ViT-B/16 and 58.0\% with ViT-B/32 among all models trained on the synthetic caption datasets (e.g., CC3M, CC12M, YFCC15M, and Merged-30M). However, the same models trained on PixelProse do not achieve the best accuracy in retrieval tasks according to Tab.\ 1 main and \cref{tab:zs_ret_vit32}. We infer that this might be due to the length and content of the captions used during the training. DreamLIP utilizes three MLLMs to generate long synthetic captions, resulting in longer captions than those generated by PixelProse, which only uses Gemini Pro. Consequently, models trained on longer and more varied captions are better at capturing local information (i.e., better at retrieval task), while models trained on relatively shorter and similar captions are better at focusing on global objects in the image (i.e., better at classification task). Investigating this trade-off based on the nature of these long captions would be an intriguing direction for future work.

\begin{table*}[h!]
    \setlength\tabcolsep{1.0pt}
    \centering
    \resizebox{0.73\linewidth}{!}{
    \begin{tabular}{c<{\hspace{3.0em}}c<{\hspace{1.0em}}|c<{\hspace{0.7em}}c<{\hspace{0.7em}}c<{\hspace{0.7em}}c<{\hspace{0.7em}}c<{\hspace{0.7em}}c<{\hspace{0.7em}}c<{\hspace{0.7em}}c<{\hspace{0.7em}}c<{\hspace{0.7em}}c<{\hspace{0.7em}}c<{\hspace{0.7em}}|c<{\hspace{0.7em}}}
        \toprule[1.2pt]
        Data&Model&
        \rotatebox[origin=lb]{90}{\smash{ Food-101}} & \rotatebox[origin=lb]{90}{\smash{ CIFAR-10}} & \rotatebox[origin=lb]{90}{\smash{ CIFAR-100}}   & \rotatebox[origin=lb]{90}{\smash{ SUN397}}   & \rotatebox[origin=lb]{90}{\smash{ Cars}}     & \rotatebox[origin=lb]{90}{\smash{ Aircraft}}    & \rotatebox[origin=lb]{90}{\smash{ DTD}}      & \rotatebox[origin=lb]{90}{\smash{ Pets}}     & \rotatebox[origin=lb]{90}{\smash{ Caltech-101}} &
        \rotatebox[origin=lb]{90}{\smash{ Flowers}}  & \rotatebox[origin=lb]{90}{\smash{ ImageNet}}  & \rotatebox[origin=lb]{90}{\smash{ Average}} \\
        \midrule
        \multirow{4}{3em}{\rotatebox[origin=c]{0}{CC3M}} 
        & CLIP~\cite{radford2021learning}  & 14.2 & 73.5 & 37.8 & 39.3 & 2.4 & 1.4 & 14.4 & 16.9 & 66.4 & 10.7 & 19.7 & 27.0 \\
        & SigLIP~\cite{zhai2023sigmoid}    & 13.5  & 74.2 & 40.1 & 39.8 & 2.8 & 1.3 & 12.9 & 16.3 & 67.3 & 11.3 & 21.2 & 27.3 \\
        & DreamLIP$^{*}$~\cite{zheng2025dreamlip} & 16.1& \textbf{82.0}& 45.4& 41.3& 2.5& 1.0& 13.9& 18.8& 64.4& 14.1& 25.9 & 29.6\\
        & \rc\method                              & \rc\textbf{24.9} & \rc80.0 & \rc\textbf{51.9} & \rc\textbf{51.5} & \rc\textbf{3.8} & \rc\textbf{1.5} & \rc\textbf{25.2} & \rc\textbf{29.9} & \rc\textbf{77.2} & \rc\textbf{21.9} & \rc\textbf{33.0} & \rc\textbf{36.4} \\ 
        \midrule
        \multirow{4}{3em}{\rotatebox[origin=c]{0}{CC12M}} 
        & CLIP~\cite{radford2021learning}  & 41.6 & 88.5 & 57.5 & 54.5 & 11.3 & 2.3 & 23.4 & 43.3 & 80.4 & 15.7 & 37.4 & 41.4 \\
        & SigLIP~\cite{zhai2023sigmoid}    & 41.9 & 87.5 & 59.3 & 55.2 & 11.8 & 1.6 & 27.3 & 41.8 & 80.9 & 18.4 & 37.8 & 42.1 \\
        & DreamLIP$^{*}$~\cite{zheng2025dreamlip} & 48.9& 86.4& 63.0& 55.7& 17.9& 1.9& 23.5& 41.9& 83.2& 25.8& 44.2 & 44.8     \\ 
        & \rc\method                         & \rc\textbf{52.9} & \rc\textbf{91.2} & \rc\textbf{67.5} & \rc\textbf{61.0} & \rc\textbf{23.8} & \rc\textbf{3.7} & \rc\textbf{32.1} & \rc\textbf{54.2} & \rc\textbf{85.5} & \rc\textbf{30.6} & \rc\textbf{46.7} & \rc\textbf{49.9} \\ 
        \midrule
        \multirow{4}{3em}{\rotatebox[origin=c]{0}{YFCC15M}} 
        & CLIP~\cite{radford2021learning}  & 38.9 & 86.2 & 58.2 & 53.3 & 7.1 & 4.0 & 23.9 & 27.6 & 76.8 & 38.0 & 38.9 & 41.2 \\
        & SigLIP~\cite{zhai2023sigmoid}    & 37.7 & 86.1 & 57.1 & 53.2 & 6.4 & 4.3 & 25.3 & 30.4 & 77.4 & 35.3 & 38.6 & 41.1 \\
        & DreamLIP$^{*}$~\cite{zheng2025dreamlip} & \textbf{51.7} & \textbf{87.9} & \textbf{60.7} & \textbf{54.8} & 9.4 & \textbf{7.1} & 26.8 & 36.3 & 79.6 & \textbf{48.6} & 46.6 & \textbf{46.3}  \\ 
        & \rc\method                     & \rc40.3 & \rc84.9 & \rc57.0 & \rc54.6 & \rc\textbf{13.5} & \rc5.9 & \rc\textbf{31.3} & \rc\textbf{38.6} & \rc\textbf{82.1} & \rc47.8 & \rc\textbf{48.1} & \rc45.8 \\ 
        \midrule
        \multirow{4}{3em}{\rotatebox[origin=c]{0}{Merged-30M}}  
        & CLIP~\cite{radford2021learning}  & 54.8 & 90.0 & 67.1 & 62.0 & 13.0 & 3.6 & 27.6 & 49.4 & 83.8 & 41.4 & 45.6 & 48.9\\
        & SigLIP~\cite{zhai2023sigmoid}    & 52.8 & 90.8 & 66.1 & 63.4 & 15.0 & 5.4 & 29.9 & 47.8 & 84.4 & 35.7 & 46.5 & 48.9 \\
        & DreamLIP$^{*}$~\cite{zheng2025dreamlip}  & \textbf{68.2} & \textbf{91.8} & 69.2 & 62.2 & 20.7 & \textbf{8.0} & 32.1 & \textbf{62.8} & \textbf{86.1} & 48.5 & \textbf{55.7} & 55.0  \\ 
        & \rc\method           & \rc65.9 & \rc91.5 & \rc\textbf{70.8} & \rc\textbf{64.6} & \rc\textbf{23.4}& \rc7.6 & \rc\textbf{37.6} & \rc57.3 & \rc\textbf{86.1} & \rc\textbf{52.2} & \rc53.4 & \rc\textbf{55.5} \\ 
        \midrule
        \multirow{1}{3em}{\rotatebox[origin=c]{0}{LAION-400M}}  
        &  OpenCLIP$^{\dag}$~\cite{cherti2023reproducible} & 78.2 & 88.4 & 68.3 & 65.0 & 74.5 & 14.6 & 52.4 & 84.9 & 88.4 & 65.9 & 60.2 & 67.3 \\
        \midrule
        \multirow{1}{3em}{\rotatebox[origin=c]{0}{{DataComp-1B}}}  
        &  OpenCLIP$^{\dag}$~\cite{cherti2023reproducible} & 86.3 & 95.6 & 80.4 & 67.3 & 87.3 & 24.8 & 57.2 & 90.2 & 91.6 & 73.2 & 69.2 & 74.8 \\
        \midrule
        \multirow{1}{3em}{\rotatebox[origin=c]{0}{{LAION-2B}}}  
        &  OpenCLIP$^{\dag}$~\cite{cherti2023reproducible} & 82.7 & 93.6 & 75.8 & 68.7 & 86.1 & 24.5 & 55.8 & 90.4 & 90.5 & 71.6 & 66.5 & 73.3  \\\midrule
        \multirow{2}{3em}{\rotatebox[origin=c]{0}{{2.5B}}}  
        &  MetaCLIP$^{*}$~\cite{xu2023demystifying} & 82.7 & 95.2 & 77.7 & 66.8 & 77.4 & 27.0 & 58.9 & 90.9 & 92.8 & 69.9 & 67.6 & 73.4   \\
        &  Llip$^{*}$~\cite{lavoie2024modeling} & 84.1 & 95.5 & 80.8 & 68.6 & 82.2 & 34.9 & 58.8 & 92.3 & 92.9 & 74.8 & 67.5 & 75.6\\
        \bottomrule[1.2pt]
    \end{tabular}}
    \vspace{-2mm}
    \caption{\textbf{Zero-shot classification results} in terms of top-1 accuracy on the ImageNet~\cite{deng2009imagenet} and Food101~\cite{bossard2014food}, CIFAR-10~\cite{krizhevsky2009learning}, CIFAR-100~\cite{krizhevsky2009learning}, SUN397~\cite{xiao2010sun}, Stanford Cars~\cite{KrauseStarkDengFei-Fei_3DRR2013}, FGVCAircraft~\cite{maji2013fine}, DTD~\cite{cimpoi2014describing}, Oxford Pets~\cite{parkhi2012cats}, Caltech101~\cite{fei2004learning}, Flowers102~\cite{nilsback2008automated}, and ImageNet~\cite{deng2009imagenet} datasets. The vision encoder architecture is ViT-B/32. The best results are highlighted in \textbf{bold}. Results are reproduced with our setup for fair comparison unless otherwise marked. $^{*}$: Results copied from their work. $^{\dag}$: Results obtained using their official pre-trained weights.}
    \label{tab:zs_cla_vit32}
\end{table*}

\begin{table*}[t]
\centering
\setlength\tabcolsep{1.0pt}
\resizebox{.75\linewidth}{!}{

\begin{tabular}
{c<{\hspace{1.0em}} | c<{\hspace{0.7em}}c<{\hspace{0.7em}}c<{\hspace{0.7em}} | c<{\hspace{0.7em}}c<{\hspace{0.7em}}c<{\hspace{0.7em}} | c<{\hspace{0.7em}}c<{\hspace{0.7em}}c<{\hspace{0.7em}} | c<{\hspace{0.7em}}c<{\hspace{0.7em}}c<{\hspace{0.7em}}}

\toprule
&\multicolumn{6}{c|}{Flickr30K}& \multicolumn{6}{c}{MSCOCO}\\
&\multicolumn{3}{c|}{Image-to-text} & \multicolumn{3}{c|}{Text-to-image} & \multicolumn{3}{c|}{Image-to-text} & \multicolumn{3}{c}{Text-to-image} \\
Method & R@1 & R@5 & R@10 & R@1 & R@5 & R@10 & R@1 & R@5 & R@10 & R@1 & R@5 & R@10 \\
\midrule
\multicolumn{13}{c}{ \textit{Model Architecture: ViT-B/32}}\\
\midrule
CLIP~\cite{radford2021learning} & 76.1 & 92.1 & 95.7 & 57.5 & 81.8 & 88.6 & 49.4 & 73.7 & 82.2 & 32.0 & 57.1 & 68.2  \\ 
SigLIP~\cite{zhai2023sigmoid}  & 74.4 & 92.2 & 96.1 & 56.8 & 81.3 & 87.7 & 48.9 & 74.4 & 82.2 & 32.0 & 57.2 & 68.0 \\ 
\rc\method                   & \rc\textbf{85.6} & \rc\textbf{96.9} & \rc\textbf{98.3} & \rc\textbf{66.3} & \rc\textbf{87.6} & \rc\textbf{92.7} & \rc\textbf{57.2} & \rc\textbf{80.9} & \rc\textbf{88.0} & \rc\textbf{38.9} & \rc\textbf{64.8} & \rc\textbf{74.6}  \\ 
\midrule
\multicolumn{13}{c}{ \textit{Model Architecture: ViT-B/16}}\\
\midrule
CLIP~\cite{radford2021learning} & 85.2 & 96.2 & 98.6 & 66.3 & 87.8 & 93.2 & 56.9 & 79.7 & 87.4 & 38.0 & 64.1 & 74.5  \\ 
SigLIP~\cite{zhai2023sigmoid}  & 85.4 & 96.7 & 97.8 & 66.6 & 88.4 & 93.5 & 57.5 & 80.5 & 87.6 & 38.9 & 64.5 & 74.4 \\ 
\rc\method                           & \rc\textbf{89.9} & \rc\textbf{98.5} & \rc\textbf{99.5} & \rc\textbf{73.6} & \rc\textbf{92.0} & \rc\textbf{95.4} & \rc\textbf{62.4} & \rc\textbf{84.0} & \rc\textbf{89.7} & \rc\textbf{43.4} & \rc\textbf{69.3} & \rc\textbf{78.8}  \\ 
\bottomrule         
\end{tabular}}
    \caption{\textbf{PixelProse: Zero-shot image-text retrieval results} in terms of R@1, R@5, and R@10 on the Flickr30K~\cite{young2014image} and MSCOCO~\cite{lin2014microsoft} datasets. The best results are highlighted in \textbf{bold}. Results are reproduced with our setup for fair comparison.
    }
\label{tab:pixelprose}
\end{table*}

\begin{table*}[h]
    \setlength\tabcolsep{1.0pt}
    \centering
    \resizebox{0.65\linewidth}{!}{
    \begin{tabular}{c<{\hspace{1.0em}}|c<{\hspace{0.7em}}c<{\hspace{0.7em}}c<{\hspace{0.7em}}c<{\hspace{0.7em}}c<{\hspace{0.7em}}c<{\hspace{0.7em}}c<{\hspace{0.7em}}c<{\hspace{0.7em}}c<{\hspace{0.7em}}c<{\hspace{0.7em}}c<{\hspace{0.7em}}|c<{\hspace{0.7em}}}
        \toprule[1.2pt]
        Model&
        \rotatebox[origin=lb]{90}{\smash{ Food-101}} & \rotatebox[origin=lb]{90}{\smash{ CIFAR-10}} & \rotatebox[origin=lb]{90}{\smash{ CIFAR-100}}   & \rotatebox[origin=lb]{90}{\smash{ SUN397}}   & \rotatebox[origin=lb]{90}{\smash{ Cars}}     & \rotatebox[origin=lb]{90}{\smash{ Aircraft}}    & \rotatebox[origin=lb]{90}{\smash{ DTD}}      & \rotatebox[origin=lb]{90}{\smash{ Pets}}     & \rotatebox[origin=lb]{90}{\smash{ Caltech-101}} &
        \rotatebox[origin=lb]{90}{\smash{ Flowers}}  & \rotatebox[origin=lb]{90}{\smash{ ImageNet}}  & \rotatebox[origin=lb]{90}{\smash{ Average}} \\
        \midrule
        \multicolumn{13}{c}{ \textit{Model Architecture: ViT-B/32}}\\
        \midrule
        CLIP~\cite{radford2021learning}   & 52.4 & 85.2 & 56.1 & 54.1 & 26.0 & 1.9 & 32.0 & 59.0 & 85.0 & 34.4 & 44.9 & 48.3  \\
        SigLIP~\cite{zhai2023sigmoid}    & 53.8 & 85.7 & 59.4 & 55.6 & 25.2 & 3.1 & 32.1 & 64.1 & 84.8 & 37.1 & 44.6 & 49.6 \\
        \rc\method                          & \rc\textbf{66.3} & \rc\textbf{89.9} & \rc\textbf{67.4} & \rc\textbf{60.4} & \rc\textbf{47.0} & \rc\textbf{4.4} & \rc\textbf{37.3} & \rc\textbf{76.6} & \rc\textbf{89.5} & \rc\textbf{45.1} & \rc\textbf{54.3} & \rc\textbf{58.0} \\ 
        \midrule
        \multicolumn{13}{c}{ \textit{Model Architecture: ViT-B/16}}\\
        \midrule
        CLIP~\cite{radford2021learning}   & 63.2 & 87.9 & 60.2 & 59.1 & 37.0 & 2.3 & 40.1 & 68.8 & 86.6 & 42.1 & 50.8 & 54.4  \\
        SigLIP~\cite{zhai2023sigmoid}    & 63.4 & 87.5 & 58.6 & 60.5 & 36.2 & 3.5 & 37.8 & 63.5 & 87.0 & 43.4 & 51.1 & 53.9 \\
        \rc\method                          & \rc\textbf{72.1} & \rc\textbf{92.1} & \rc\textbf{71.9} & \rc\textbf{60.8} & \rc\textbf{53.0} & \rc\textbf{3.6} & \rc\textbf{43.4} & \rc\textbf{74.4} & \rc\textbf{90.1} & \rc\textbf{46.8} & \rc\textbf{59.6} & \rc\textbf{60.7}\\ 
        \bottomrule[1.2pt]
    \end{tabular}}
    \caption{\textbf{PixelProse: Zero-shot classification results} in terms of top-1 accuracy on ImageNet~\cite{deng2009imagenet} and 10 common downstream datasets. The best results are highlighted in \textbf{bold}. Results are reproduced with our setup for fair comparison.
    }
    \label{tab:pixelprose2}
\end{table*}

\begin{table*}[t]
    \centering
    \begin{tabular}{llcccccccc}
    \toprule
    &\multirow{2}{*}{\textbf{Method}} & \multicolumn{1}{c}{ImageNet} & \multicolumn{2}{c}{MSCOCO} & \multicolumn{2}{c}{Flickr30k} & \multicolumn{2}{c}{Training} \\\cmidrule(lr){3-3}\cmidrule(lr){4-5}\cmidrule(lr){6-7} \cmidrule(lr){8-9} 
                  &              & Top-1 & I2T@1 & T2I@1 & I2T@1 & T2I@1 & Time & Mem.\\\midrule
    \rownumber{1} & CLIP: 1 globals + 0 locals    & 24.5 & 38.9  & 26.8  & 69.1  & 51.2  & 3.8h & 10.5G   \\
     \rownumber{2} &  CLIP: 2 globals + 0 locals    & 29.7 & 47.6  & 33.8  & 78.1  & 60.7  & 12.2h & 16.3G   \\
     \rownumber{3} & CLIP: 2 globals + 2 locals   & 31.3 & 49.6  & 35.9  & 78.4  & 62.1  & 17.6h & 20.6G   \\
     \rownumber{4} &  \method: 2 globals + 0 locals        & 31.0 & 50.1  & 35.4  & 80.9  & 63.2  & 14.6h & 18.2G   \\
     \rownumber{5} & \method: 2 globals + 2 locals        & 34.8 & 51.5  & 38.5  & 82.7  & 65.9  &  21.4h & 23.4G \\
     \rownumber{6} & \method: 2 globals + 4 locals        & 36.3 & 52.3  & 39.6  & 83.5  & 67.4  & 26.5h & 28.2G \\
     \rownumber{7} & \method: 2 globals + 6 locals        & 37.1 & 53.1  & 40.1  & 84.1  & 68.6  & 31.7h & 32.6G \\
    \bottomrule
    \end{tabular}
    \caption{\textbf{Ablation on the training efficiency of \method.} Total running time and peak memory usage per GPU are shown with different configuration of global and locap crops. We also report zero-shot classification results on ImageNet~\cite{deng2009imagenet} and zero-shot retrieval results on Flickr30K~\cite{young2014image} and MSCOCO~\cite{lin2014microsoft}. We train our model (ViT-B/16) on four 4-GPU machines using the CC3M dataset with the batch size of 1,024.}
    \label{tab:ablation2}
\end{table*}

\subsection{Ablation Study on Training Efficiency}

In \cref{tab:ablation2}, we report the GPU memory requirements and training time based on the number of global and local crops, along with the corresponding zero-shot performance. Compared to CLIP, \method requires 10-20\% more training time and GPU memory due to its teacher-student setup (row~\rownumber{2} vs row~\rownumber{4}, and row~\rownumber{3} vs row~\rownumber{5}). As no gradients flow into the teacher during training, \method does not add too much computational overhead while outperforming CLIP.

For \method, we fixed the number of global crops for both image and text to two, while varying the number of local crops included during training (row~\rownumber{4}-row~\rownumber{7}). As a result, increasing the number of local crops improves zero-shot classification and retrieval tasks, achieving 37.1\% accuracy on ImageNet and 53.1\% and 40.1\% R@1 scores on image-to-text and text-to-image retrieval on the MSCOCO validation set with six local crops. The improvement is most significant between zero and two local crops (row~\rownumber{4} vs row~\rownumber{5}), with diminishing returns as more local crops are added. GPU memory usage and training time also increase with the number of local crops. Therefore, one could determine the optimal number of local crops based on the available computational resources.

\subsection{Experiment on the Winoground Dataset}

\begin{table}[h]
    \centering
    \setlength\tabcolsep{3pt}
    \resizebox{0.7\linewidth}{!}{
    \begin{tabular}{lc|ccc}
    \toprule
    Method &  Data Size & Text & Image & Group\\\midrule
    CLIP~\cite{radford2021learning} & 30M & 29.3 & 13.0 & 8.5  \\ 
    SigLIP~\cite{zhai2023sigmoid}   & 30M & 29.0 & 11.0 & 8.3  \\
    DreamLIP~\cite{zheng2025dreamlip} & 30M & 27.8 & 15.8 & \textbf{11.5} \\
    \rc\method                       & \rc30M & \rc\textbf{30.8} & \rc\textbf{16.5}  & \rc11.3 \\
    \midrule
                        & 400M & 25.5 & 11.5 & 7.8 \\
    OpenCLIP~\cite{cherti2023reproducible} & 1B & 29.8 & 8.8 & 7.3 \\
                    & 2B & 28.0 & 10.8 & 8.3 \\
    \bottomrule
    \end{tabular}}
    \caption{\textbf{Evaluation on Winoground~\cite{thrush2022winoground}.}}
    \label{tab:winoground}
\end{table}

In addition to the visual perception and contextual understanding tasks presented in the main paper, we also evaluate \method on the Winoground dataset~\cite{thrush2022winoground}. Each entry in the dataset consists of two images and two captions, where the task is to correctly match the image-text pairs (25\% by random chance). Both captions contain an identical set of words, but in different orders, assessing whether the model possesses sufficient compositional understanding for challenging image-text pairs.

In \cref{tab:winoground}, we compare \method with three baselines (CLIP~\cite{radford2021learning}, SigLIP~\cite{zhai2023sigmoid}, and DreamLIP~\cite{zheng2025dreamlip}) as well as OpenCLIP~\cite{cherti2023reproducible} models trained on larger datasets. The results of the OpenCLIP models indicate that scaling up the size of pre-training datasets does not necessarily improve the performance. In other words, large datasets alone cannot ensure high compositional knowledge in VLMs. However, \method outperforms all OpenCLIP models, suggesting that our self-distillation and cross-modality learning effectively enhance the model's contextual understanding. \method achieves 30.8\% and 16.5\% in text and image scores, respectively, surpassing the previous best baselines of 29.3\% text score from CLIP and 15.8\% image score from DreamLIP, while achieving comparable results in terms of group score.

\subsection{Experiment on MLLM setting}\label{subsec:supp_mllm}

 \begin{table}[h]
    \vspace{-1mm}
    \centering
    \setlength\tabcolsep{1.5pt}
    \scalebox{0.8}{
    \begin{tabular}{lcccccc}
    \toprule
    Method & Data Size & ScienceQA & POPE & GQA & TextVQA & MMMU  \\\midrule
    CLIP~\cite{radford2021learning} &30M    & 65.2&	80.1&	58.7&	53.5&	33.9\\
    OpenCLIP~\cite{cherti2023reproducible} &400M  & 67.2&	 81.3&	59.8&	54.4&	36.5\\
    \rc \method & \rc 30M  & \rc \textbf{67.8}& \rc	\textbf{83.2}& \rc	\textbf{60.4}&	\rc \textbf{55.3}&	 \rc \textbf{36.8}\\ 
    \bottomrule
    \end{tabular}
    }
    \caption{\textbf{LLaVA v1.5 experiment}. CLIP and \method trained on Merged-30M and OpenCLIP trained on LAION-400M.} 
    \label{tab:llava_exp}
\end{table}
\noindent
In order to evaluate \method on MLLM setting, we adapt our vision encoder (ViT-B/16) to LLaVA framework~\cite{liu2024improved}. We follow the training process of LLaVA v1.5 which consists of feature alignment pre-training and visual instruction tuning. Various benchmarks are selected for comprehensive evaluation including ScienceQA~\cite{lu2022learn}, POPE~\cite{li2023evaluating}, GQA~\cite{hudson2019gqa}, TextVQA~\cite{singh2019towards}, and MMMU~\cite{yue2024mmmu}. In \cref{tab:llava_exp}, \method outperforms both CLIP trained on the same data (Merged-30M) and OpenCLIP trained on LAION-400M, which demonstrates the effectiveness of \method on visual
reasoning and compositional question answering.

\subsection{Ablation Study on Text Cropping Strategy}

\begin{table}[h]
    \centering
    
    \resizebox{\linewidth}{!}{
    \begin{tabular}{lccccc}
    \toprule
    \multirow{2}{*}{\textbf{Method}} & \multicolumn{1}{c}{ImageNet} & \multicolumn{2}{c}{MSCOCO} & \multicolumn{2}{c}{Flickr30K} \\\cmidrule(lr){2-2}\cmidrule(lr){3-4}\cmidrule(lr){5-6}
                              & Top-1  & I2T@1 & T2I@1 & I2T@1 & T2I@1 \\\midrule
    Masked text~\cite{li2021supervision} & 28.5  & 46.0 & 32.8  & 76.2 & 59.2 \\ 
    Summarized text~\cite{gao2022pyramidclip} & 33.9 & 51.7  & 37.8 & 81.0  & 65.3 \\
    Local within global  & 34.6  &  52.1 & 37.8  &81.2 & 66.1\\
    \rc \method              & \rc\textbf{35.1}  & \rc\textbf{52.6} & \rc\textbf{38.9}  & \rc\textbf{83.0} & \rc\textbf{66.5}\\
    \bottomrule
    \end{tabular}
    }
    \caption{\textbf{Ablation on text cropping strategies.} We compare various text cropping methods in terms of zero-shot classification results on ImageNet~\cite{deng2009imagenet} and zero-shot retrieval results on Flickr30K~\cite{young2014image} and MSCOCO~\cite{lin2014microsoft}. All models are trained on CC3M with batch size 1,024 and ViT-B/16 image encoder.}
    \label{tab:ablation_textcropping}
\end{table}

In our paper, both global and local crops of captions are randomly sampled from long synthetic captions. Typically, global texts (1-5 sentences) are longer than local texts (1 sentence). The sampling processes of global and local crops are entirely independent. Therefore, global captions may or may not include local captions, similar to the image crop method in SimCLR~\cite{chen2020simple} and DINO~\cite{caron2021emerging}. While mismatches can occur between global and local captions via random sampling, as with the global and local crops of images, the model can learn conceptual correspondences. For example, for an image of a park, the global text may describe the park, while the local text describes a dog. The model then learns that "park" and "dog" are related in the text.

To validate our text cropping method, we compare various cropping strategies in \cref{tab:ablation_textcropping}, referring to previous works. In the \textit{masked text} setting, we randomly select one to five sentences from a long caption and set it as the global caption. Then, 15\% of text tokens are replaced with the $[\text{mask}]$ token, which is used as the local caption. This setting is similar to text self-supervised learning in DeCLIP~\cite{li2021supervision}. In the \textit{summarized text} setting, we sample one summary sentence as the global caption and one detailed sentence as the local caption, similar to PyramidCLIP~\cite{gao2022pyramidclip}. As the synthetic captions of DreamLIP~\cite{zheng2025dreamlip} already distinguish between short and long captions, which generally describe the summary and details respectively, we directly use their categories to construct global and local crops. In the \textit{local within global} setting, we ensure that local captions are always included in global captions, while the rest remains the same as our text cropping method. The results show that our text-cropping strategy, inspired by image cropping, performs the best, as it learns various conceptual similarities via independent random sampling.

\subsection{Additional Results on Semantic Segmentation}\label{subsec:supp_add_seg}
  \begin{table}[h]
    \centering
    \resizebox{\linewidth}{!}{
    \begin{tabular}{lccccc}
    \toprule
    Method & VOC20 & City. & Context59 & ADE & COCO-Stf. \\\midrule
    CLIP~\cite{radford2021learning}   & 11.3   &  5.0   &  4.5   &  1.3 &  2.8\\
    SigLIP~\cite{zhai2023sigmoid}   &  14.5   &  5.5   &  5.8   &  2.2   &  3.8\\
    DreamLIP~\cite{zheng2025dreamlip} & 1.8   &  0.9   &  0.4   &  0.1   &  0.1 \\ 
    \rc \method  & \rc\textbf{53.6}   & \rc\textbf{13.9}    & \rc\textbf{15.7}   & \rc\textbf{8.5}    & \rc\textbf{10.7}\\
    \bottomrule
    \end{tabular}
    }
    \caption{\textbf{Zero-shot semantic segmentation results} in terms of mean Intersection over Union (mIoU). The vision encoder architecture is ViT-B/16 and all models are trained on Merged-30M.
    } 
    \label{tab:additional_segmentation}
\end{table}

In addition to Table 3 in main paper, we compare segmentation performance of \method to CLIP~\cite{radford2021learning}, SigLIP~\cite{zhai2023sigmoid}, and DreamLIP~\cite{zheng2025dreamlip} in \cref{tab:additional_segmentation}. Surprisingly, DreamLIP performs poorly in every semantic segmentation benchmarks, likely because its loss function weakens local image representation by matching local visual patches with global text. Ours is by far the best performing method, likely due to its cross-modality embedding and local-to-global matching, alleviating local feature suppression.

\subsection{Additional SOTA Comparison}

\begin{table}[h]
        \centering
    \resizebox{\linewidth}{!}{
        \begin{tabular}{lccccccccc}
        \toprule
        \multirow{2}{*}{Method}& \multirow{2}{*}{Data Size} & \multirow{2}{*}{Batch Size} & \multicolumn{2}{c}{MSCOCO} & \multicolumn{2}{c}{Flickr30K} \\\cmidrule(lr){4-5}\cmidrule(lr){6-7}
               &   &  & I2T & T2I & I2T & T2I \\\midrule
        VeCLIP~\cite{lai2025veclip} & 300M   & 32k & 67.8 & 48.9  & 91.2 & 76.3\\
        MobileCLIP-B~\cite{vasu2024mobileclip} & 1B & 65k & \textbf{68.8} & 50.6 & 91.4 & 77.3 \\
        \rc\method & \rc30M  & \rc4k & \rc68.0 & \rc\textbf{52.5}  & \rc\textbf{92.9} & \rc\textbf{80.3}  \\ 
        \bottomrule
        \end{tabular}
        }
        \caption{\textbf{Comparison to VeCLIP~\cite{lai2025veclip} and MobileCLIP~\cite{vasu2024mobileclip}} in terms of zero-shot retrieval results on Flickr30K~\cite{young2014image} and MSCOCO~\cite{lin2014microsoft}. } 
        \label{tab:sota_comparison_veclip}
\end{table}
 
In Table 1 and 2 in the main paper, \method is compared to other methods using multi-modal data augmentation including DreamLIP~\cite{zheng2025dreamlip}, LaCLIP~\cite{fan2024improving}, and MLLM-A~\cite{liu2023mllms}. We additionally report the results of VeCLIP~\cite{lai2025veclip} and MobileCLIP~\cite{vasu2024mobileclip} for comparison. VeCLIP exploits LLaVA~\cite{liu2024visual} to generate detailed captions while MobileCLIP utilizes CoCa~\cite{yu2022coca} to generate multiple synthetic cations. Although \method is trained with much smaller pre-training set and batch size, it outperforms other methods, demonstrating the efficient usage of synthetic captions within our framework.

\subsection{Rescaling Loss in Previous Methods}

\begin{table}[h]
    \centering
    \resizebox{\linewidth}{!}{\begin{tabular}{lcccccc}
    \toprule
    \multirow{2}{*}{\textbf{Method}} & \multicolumn{1}{c}{ImageNet} & \multicolumn{2}{c}{MSCOCO} & \multicolumn{2}{c}{Flickr30k} \\\cmidrule(lr){2-2}\cmidrule(lr){3-4}\cmidrule(lr){5-6}  
                                & Top-1 & I2T@1 & T2I@1 & I2T@1 & T2I@1  \\\midrule
    CLIP w/ Aug. EMA   & 20.0 & 19.2  & 14.7  & 38.7 & 29.3   \\ \midrule                      
    SILC (1.0, 1.0) & 14.7 & 9.2  & 13.4  & 26.8 & 18.7   \\
    SILC (1.5, 0.5) & 19.0 & 17.4  & 12.9  & 33.8 & 25.2   \\
    SILC (1.8, 0.2) & 21.0 & 20.8  & 14.8  & 40.2 & 29.3   \\
    SILC (1.9, 0.1) & \textbf{21.4} & \textbf{21.1}  & \textbf{15.3}  & \textbf{42.0} & \textbf{29.7}   \\
    SILC (1.95, 0.05) & \textbf{21.4} & 20.4  & 15.1  & 40.2 & 29.5   \\
    \bottomrule
    \end{tabular}}
    \caption{\textbf{Ablation on the loss scale of SILC~\cite{naeem2023silc}.} SILC with different loss scale (a,b) where the total loss is calculated as $\mathcal{L}_{\text{total}} = a\mathcal{L}_{\text{CLIP}} + b\mathcal{L}_{\text{self-distill}}$. Models are trained on CC3M with the batch size of 1,024 and one global and one local crops are used as augmentation.}
    \label{tab:loss_scale_ablation}
\end{table}

In \cref{tab:loss_scale_ablation}, we conduct an experiment to demonstrate the effect of loss scaling in SILC~\cite{naeem2023silc}, which also utilizes self-supervision in contrastive vision-language pre-training. We adjust the scale of the CLIP loss ($a$) and the self-distillation loss ($b$), while maintaining their sum constant (i.e., $a + b = 2$). For comparison, we also include CLIP~\cite{radford2021learning} with the same augmentation and EMA. Interestingly, a naive summation of the two losses (i.e., $a = 1.0, b = 1.0$) results in a worse performance compared to CLIP, highlighting the importance of selecting appropriate scaling parameters for optimal performance. Consequently, $a = 1.9$ and $b = 0.1$ yield the best results, which we used to reproduce their results in \cref{tab:ablation3}.

Similarly, other works~\cite{li2021supervision, mu2022slip, dong2023maskclip, sameni2024building} that integrated self-supervised learning in VLM training adopt loss scaling parameters to balance the learning speed between the contrastive objective and the self-distillation objective. This is due to the different scales of the loss functions, as previous works often employ the symmetric InfoNCE loss~\cite{oord2018representation} for contrastive learning while using the cross-entropy loss between masked inputs or global and local crops for self-supervised learning. We unify the loss function with the InfoNCE loss, eliminating the need for a scaling factor, since the CLIP loss and the COSMOS loss are already updated on the same scale.


\end{document}